\documentclass{article}

\usepackage{graphicx} % more modern
\usepackage{subfigure}

\usepackage{natbib}

\usepackage{algorithm}
\usepackage{algorithmic}

\usepackage{xcolor,calc}
\usepackage{hyperref}
\usepackage[compact]{titlesec}
\usepackage{amsmath,amsthm}
\usepackage{amsopn,amssymb}
\usepackage{stmaryrd}
\usepackage{booktabs,examplep}

\usepackage[english]{babel}

\usepackage[mathscr]{euscript}
\usepackage{hyperref}

\usepackage[accepted]{icml2013}
\newtheorem{theorem}{Theorem}[section]
\newtheorem{lemma}[theorem]{Lemma}
\newtheorem{e-proposition}[theorem]{Proposition}

\newtheorem{e-definition}[theorem]{Definition\rm}

\def\ddelta{{\boldsymbol\delta}}

\definecolor{vertsombre}{rgb}{0.00,0.57,0.1}

\def\hat{\widehat}
\def\tilde{\widetilde}

\def\diag{\text{\rm diag}}
\def\sff{{\sf f}}
\def\sffstar{{\sf f}^*}
\def\sfb{{\sf b}}
\def\sfbstar{{\sf b}^*}

\def\sfsstar{{\sf s}^*}
\def\sfr{{\sf r}}
\def\sfrstar{{\sf r}^*}
\def\sfg{{\sf g}}

\def\xxi{{\boldsymbol\xi}}

\def\DDelta{{\boldsymbol\Delta}}

\def\aalpha{{\boldsymbol\alpha}}
\def\bbeta{{\boldsymbol\beta}}
\def\pphi{{\boldsymbol\phi}}

\def\llambda{{\boldsymbol\lambda}}

\def\nnu{{\boldsymbol\nu}}
\def\bfM{{\mathbf M}}
\def\bfX{{\mathbf X}}
\def\bfD{{\mathbf D}}
\def\AA{{\mathbf A}}
\def\BB{{\mathbf B}}

\def\DD{{\mathbf D}}

\def\II{{\mathbf I}}

\def\Ppi{{\boldsymbol\Pi}}

\def\bx{{\boldsymbol x}}
\def\bu{{\boldsymbol u}}

\def\bv{{\boldsymbol v}}
\def\bA{{\boldsymbol A}}

\def\bR{{\boldsymbol R}}
\def\bX{{\boldsymbol X}}
\def\bY{{\boldsymbol Y}}

\def\RR{\mathbb R}
\def\Ex{\mathbf E}
\def\Pb{\mathbf P}
\def\Var{\mathbf{Var}}

\def\bfR{\mathbf R}

\def\mcB{\mathcal B}

\def\mcK{\mathcal K}
\def\mcT{\mathcal T}

\def\sparsity{\text{\tiny${\mathscr{S}}$}}
\def\og{\leavevmode\raise.3ex\hbox{$\scriptscriptstyle\langle\!\langle$~}}
\def\fg{\leavevmode\raise.3ex\hbox{~$\!\scriptscriptstyle\,\rangle\!\rangle$}}
\def\1{\mathbf 1}

\def\matnorm#1{|\!|\!|#1|\!|\!|}

\newcommand{\lcf}{{\em cf.~}}

\icmltitlerunning{Learning Heteroscedastic Models by Convex Programming under Group Sparsity}

\begin{document}

\twocolumn[
\icmltitle{Learning Heteroscedastic Models by Convex Programming under Group Sparsity}

% unless you've provided the [accepted] option to the icml2013

\icmlauthor{Arnak S.~Dalalyan}{arnak.dalalyan@ensae.fr}
\icmladdress{ENSAE-CREST-GENES}
\icmlauthor{Mohamed Hebiri}{mohamed.hebiri@univ-mlv.fr}
\icmladdress{Universit\'e Paris Est}
\icmlauthor{Katia Meziani}{meziani@ceremade.dauphine.fr}
\icmladdress{Universit\'e Paris Dauphine}
\icmlauthor{Joseph Salmon}{joseph.salmon@telecom-paristech.fr}
\icmladdress{LTCI, Telecom ParisTech}

\icmlkeywords{group sparsity, heteroscedastic model, time series}

%\vskip 0.3in
\vskip -0.05in
]

\begin{abstract}
Popular sparse estimation methods based on  $\ell_1$-relaxation, such as the Lasso and the Dantzig selector,
require the knowledge of the variance of the noise in order to properly tune the regularization
parameter. This constitutes a major obstacle in applying these methods in several frameworks---such as
time series, random fields, inverse problems---for which the noise is rarely homoscedastic and its
level is hard to know in advance. In this paper, we propose a new approach to the joint estimation of
the conditional mean and the conditional variance in a high-dimensional (auto-) regression setting. An
attractive feature of the proposed estimator is that it is efficiently computable even for very large
scale problems by solving a second-order cone program (SOCP). We present theoretical analysis and
numerical results assessing the performance of the proposed procedure.
\end{abstract}

~

\vspace{-30pt}
% main text
\section{Introduction}
\label{sec:1}

% Sparse estimation methods based on $\ell_1$ relaxation, such as the Lasso and  the Dantzig selector, are powerful tools for
% estimating high dimensional linear models.

Over the last fifteen years, sparse
estimation methods based on $\ell_1$-relaxation,
among which the Lasso \cite{Tibsh} and the Dantzig selector \cite{Candes06} are the most famous examples,
have become a popular tool for estimating high dimensional linear models.
So far,  their wider use in several fields of applications (\textit{e.g.}, finance
and econometrics)  has been constrained by the difficulty of adapting to heteroscedasticity,
\textit{i.e.}, when the noise level varies across the components of the signal.
% adaptation to the case of heteroscedastic errors. The goal
% of the present work is to address this issue by extending the Dantzig selector to the model of sparse linear regression with variable
% (unknown) noise level.

Let $\mcT$ be a finite set of cardinality $T$. For every $t\in \mcT$ we observe a sequence $(\bx_t,y_t)\in\RR^d\times\RR$
obeying:
\begin{equation}\label{eq:1}
y_t=\sfbstar(\bx_t)+\sfsstar(\bx_t)\xi_t,
\end{equation}
where $\sfbstar:\RR^d\to\RR$ and ${\sfsstar}^2:\RR^d\to\RR_+$ are respectively the unknown
conditional mean and conditional
variance\footnote{
%Note that this
This formulation of the problem includes  ``time-dependent'' mean and variance, \textit{i.e.},
the case of $\Ex[y_t|\bx_t]=\sfbstar_t(\bx_t)$ and $\Var[y_t|\bx_t]=\sfsstar_t(\bx_t)$, since it is sufficient then to
consider as explanatory variable $[t;\bx_t^\top]^\top$ instead of $\bx_t$.} of $y_t$ given $\bx_t$. Then, the errors
$\xi_t$ satisfy $\Ex[\xi_t|\bx_t]=0$ and $\Var[\xi_t|\bx_t]=1$. Depending on the targeted applications,
elements of $\mcT$ may be time instances (financial engineering), pixels or voxels (image and video processing)
or spatial coordinates (astronomy, communication networks).

In this general formulation, the problem of estimating unknown functions $\sfbstar$ and $\sfsstar$ is ill-posed:
the dimensionality of unknowns is too large as compared to the number of equations  $T$,  therefore, the model is
unidentifiable. To cope with this issue, the parameters  $(\sfbstar,\sfsstar)$ are often constrained to belong to
low dimensional spaces. For instance, a common assumption is that for some given dictionary $\sff_1,\ldots,\sff_p$ of functions
from $\RR^d$ to $\RR$ and for an unknown vector $(\bbeta^*,\sigma^*)\in\RR^p\times\RR$, the relations
$\sfbstar(\bx)=[\sff_1(\bx),\ldots,\sff_p(\bx)]\bbeta^*$ and $\sfsstar(\bx)\equiv \sigma^*$ hold for every $\bx$. Even for very large
values of $p$, much larger than the sample size $T$, such a model can be efficiently learned in the sparsity scenario
using recently introduced scaled versions of \hbox{$\ell_1$-relaxations}: the  square-root Lasso \cite{Antoniadis,Belloni,Sun,Gautier}, the scaled Lasso
\cite{Stadler} and the scaled Dantzig selector \cite{DalChen}. These methods are tailored to the context of a fixed noise level across
observations (homoscedasticity),
which reduces their attractiveness for applications in the aforementioned fields.
% such as financial time series
% modeling, image and video processing, etc.
In the present work, we propose a new method of estimation for model (\ref{eq:1})
that has the appealing properties of requiring neither homoscedasticity nor any prior knowledge of the noise level.
% does not require neither noise homoscedasticity nor the knowledge of the noise level.
The only restriction we impose is that the variance function ${\sfsstar}^2$ is of reduced dimensionality,
which in our terms means that its inverse $1/\sfsstar$ is of a linear parametric form.

\paragraph{Our contributions and related work}

We propose a principled approach to the problem of joint estimation of the conditional mean function $\sfbstar$ and
the conditional variance ${\sfsstar}^2$, which boils down to a second-order cone programming (SOCP) problem.
We refer to our procedure as the Scaled Heteroscedastic Dantzig selector (ScHeDs) since it can be seen an
extension of the Dantzig selector to the case of heteroscedastic noise and group sparsity.  Note that
so far, inference under group-sparsity pioneered by \cite{Yuan_Lin_2006,Lin_2006}, has only focused on the simple case of known and
constant noise level both in the early references \cite{Nardi08,Bach_2008,Hebiri_2008,Meier_2009}, %. Some early references on the group Lasso \cite{Nardi08,Meier_2008,Bach_2008,Hebiri_2008},%\cite{Nardi08,Meier_2008,Bach_2008,Hebiri_2008,Meier_2009},
and in the more recent ones \cite{Lounici_2011, Huang_2012}. In this work we provide a theoretical analysis and
some numerical experiments assessing the quality of the proposed ScHeDs procedure.

More recently, regression estimation under the combination of sparsity and heteroscedasticity was addressed by
\cite{Daye12,Dette12,Kolar12}. Because of the inherent nonconvexity of the penalized (pseudo-)log-likelihood considered
in these works, the methods proposed therein do not estimate the conditional mean and the variance in a joint manner.
%They use an estimator of the conditional mean to estimate the conditional variance and vice-versa.
They rather rely on iterative estimation of those quantities: they alternate between the two variables,
estimating one while keeping the other one fixed.
Furthermore, the theoretical results of these papers are asymptotic. In contrast, we propose a method
that estimates the conditional mean and the variance by solving a jointly convex minimization problem and derive
nonasymptotic risk bounds for the proposed estimators.

\paragraph{Notation} We use boldface letters to denote vectors and matrices. For an integer $d>0$, we set
$[d]=\{1,\ldots,d\}$. If $\bv\in\RR^d$ and $J\subset[d]$, then $\bv_J$ denotes the sub-vector of $\bv$ obtained
by removing all the coordinates having indexes outside $J$. If $J=\{j\}$, we write $\bv_J=v_j$.
The $\ell_q$-norms of $\bv$ are defined by:
\begin{align*}
|\bv|_0&\textstyle =\sum_{j=1}^d\nolimits \1(v_j\neq 0),\quad |\bv|_\infty=\max_{j\in\{1,\ldots,d\}} |v_j|,\\
|\bv|_q^q&\textstyle=\sum_{j=1}^d\nolimits |v_j|^q,\ 1\le q<\infty.
\end{align*}
For a matrix  $\AA$, $\bA_{i,:}$ and $\bA_{:,j}$ stand respectively for its $i$-th
row and its $j$-th column.
% ;$a_{ij}$ stands for the $j$-th element of the vector $\bA_{i,:}$.
For a vector $\bY=[y_1,\ldots,y_T]^\top \in \RR^T$, we define $\text{diag}(\bY)$ as the $T\times T$
diagonal matrix having the entries of $\bY$ on its main diagonal.
%
%\paragraph{Organization of the paper}
%~\\
%
%\boxed{
%\vbox to 80pt{\hbox to 0.97\textwidth{\textbf{\color{red} TO BE COMPLETED}\hfill}}
%}

\section{Background and assumptions}

% Throughout this work,
%Without loss of generality, we assume that $\mcT=\{1,\ldots,T\}$.
We start by reparameterizing the problem as follows:
%set Then, to ease our assumptions formulation, we rescale the true signal
%by the inverse of the volatility $\sfsstar$, so that we express our conditions, not in terms of $(\sfbstar,\sfsstar)$
%but rather in terms of $(\sfrstar,\sffstar)$ defined as follows:
\begin{equation}\label{param}
\sfrstar(\bx)=1/{\sfsstar(\bx)},\qquad \sffstar(\bx)={\sfbstar(\bx)}/{\sfsstar(\bx)}.
\end{equation}
Clearly, under the condition that $\sfsstar$ is bounded away from zero, the
mapping $(\sfsstar,\sfbstar)\mapsto(\sfrstar,\sffstar)$ is bijective. As shown later,
learning the pair $(\sfrstar,\sffstar)$ appears to be more convenient than learning the original
mean-variance pair, in the sense that it can be performed by solving a convex problem.

% Instead of imposing assumptions on the pair $(\sfbstar,\sfsstar)$, we will do that on $(\sffstar,\sfrstar)$.
We  now  introduce two assumptions underlying our approach. The first
one is a group sparsity assumption on the underlying function $\sffstar$. It states that there exists a given dictionary
of functions $\sff_1,\ldots,\sff_p$ from $\RR^d$ to $\RR$ such that
$\sffstar$ is well approximated by a linear combination
$\sum_{j=1}^p\phi_j^*\sff_j$ with a (fixed) group-sparse vector $\pphi^*=[\phi_1^*,\ldots,\phi_p^*]^\top$. The
precise formulation is:
\begin{description}
\item[\textbf{Assumption (A1)}] We denote by $\bfX$ the $T\times p$ matrix having $[\sff_1(\bx_t),\ldots,\sff_p(\bx_t)]$ as $t$-th row.
Then, for a given partition $G_1,\ldots,G_K$ of $\{1,\ldots,p\}$, there is a vector $\pphi^*\in\RR^p$
such that
$[\hspace{1pt}\sffstar(\bx_1),\ldots,\sffstar(\bx_T)]^\top \approx \bfX\pphi^*$ and
$\text{Card}(\{k:{|\pphi^*_{G_k}|}_2\neq0\})\ll K$.
\end{description}

Assumption (A1) is a restriction on $\sffstar$ only; the function $\sfrstar$ does not appear in its formulation.
%Note also that the groups $G_1,\ldots,G_K$ do not necessarily form a partition of $\{1,\ldots,p\}$. In fact, in
%some situations, a few coefficients in $\pphi$ need not be penalized since the corresponding
%covariates (columns of $\bfX$) are very likely relevant for explaining the response.
Let us describe two practical situations which fit into
the framework delineated by Assumption (A1), some other examples can be found in \cite{Lounici_2011,Mairal_2011,Huang_2012}.

\paragraph{Sparse linear model with qualitative covariates}  Consider the case of linear regression with  a large number of
covariates, an important portion of which are qualitative.  Each qualitative covariate having $m$ modalities is then transformed
into a group of $m$ binary quantitative covariates. Therefore, the irrelevance of one qualitative covariate implies the
irrelevance of a group of quantitative covariates, leading to the group-sparsity condition.
% Assuming that most qualitative covariates are irrelevant for
% explaining the response, we arrive at the framework of group-sparsity.
%When most qualitative covariates are irrelevant for explaining the response, the group-sparsity assumption we introduced holds true.

\paragraph{Sparse additive model} \cite{Ravikumar_2009,Koltchinskii_Yuan10,Raskutti_et_al11}
If $\sffstar$ is a nonlinear function of a moderately
large number of quantitative covariates, then---to alleviate the curse of dimensionality---a sparse additive model
is often considered for fitting the response. This means that $\sffstar$ is assumed to
be of the simple form $\sffstar_1(x_1)+\ldots+\sffstar_d(x_d)$, with most functions $\sffstar_j$ being identically equal to zero. Projecting
each of these functions onto a fixed number of elements of a basis, $\sffstar_j(x)\approx \sum_{\ell=1}^{K_j} \phi_{\ell,j} \psi_\ell(x)$,
we get a linear formulation in terms of the unknown vector $\pphi=(\phi_{\ell,j})$. The sparsity of the additive representation implies
the group-sparsity of the vector $\pphi$.

Our second assumption requires that there is a linear space %of $\mcF$
of dimension $q$, much smaller than the sample size $T$, that contains the function $\sfrstar$. More precisely:
\begin{description}
\item[\textbf{Assumption (A2)}]  %Let $\mcF$ be the linear space of all measurable functions from $\RR^d$ to $\RR$.
%
% We are given
% There exist  .
% More precisely,
For $q$ given functions $\sfr_1,\ldots,\sfr_q$ mapping $\RR^d$ into $\RR_+$,
there is a vector $\aalpha\in\RR^q$ such that $\sfrstar(\bx)=\sum_{\ell=1}^q \alpha_\ell \sfr_\ell(\bx)$
for every $\bx\in\RR^d$.
\end{description}

%Let $\mcR$ be the linear span of $\sfr_1,\dots,\sfr_q$.
Here are two examples of functions $\sfrstar$ satisfying this assumption.

\paragraph{Blockwise homoscedastic noise} In time series modeling, one can assume that
the variance of the innovations varies smoothly over time, and, therefore, can be well approximated
by a piecewise constant function.
% The situation is similar in image processing, where it is reasonable to assume
% that neighboring pixels are corrupted by noise of nearly the same magnitude.
This situation also arises in image processing where neighboring pixels are often corrupted
by noise of similar magnitude. This corresponds to choosing a partition of $\mcT$ into $q$ cells
and to defining each $\sfr_\ell$ as the indicator function of one cell of the partition.
%$\mcR$ the set of vectors
%constant by blocks on a fixed partition of $\mcT$, \textit{i.e.}, considering the blockwise
%homoscedastic noise.

\paragraph{Periodic noise-level}
In meteorology or image processing, observations may be contaminated by a periodic noise.
In meteorology, this can be caused by seasonal variations,
whereas in image processing, this may occur if the imaging system is subject
to electronic disturbance of repeating nature. Periodic noise can be handled
by (A2) stating that $\sfrstar$  belongs to the linear span of a few trigonometric functions.

There are essentially three methods in the literature providing estimators of $(\sfbstar,\sfsstar)$ in a context
close to the one described above. All of them assume that  $\sfsstar$ is constant and equal to $\sigma^*$ and
$\left[\hspace{1pt}\sfbstar(\bx_1),\ldots,\sfbstar(\bx_T)\right]^\top=\bfX\bbeta^*$ with some sparse vector $\bbeta^*\in\RR^p$. The first
method, termed the scaled Lasso \cite{Stadler}, suggests to recover $(\bbeta^*,\sigma^*)$ by computing a solution
$(\widehat\bbeta{}^{\text{Sc-L}},\widehat\sigma^{\text{Sc-L}})$ to the optimization
problem
\begin{equation}\label{eq:2}
\min_{\bbeta,\sigma} \Big\{ T\log(\sigma)+\frac{|\bY-\bfX\bbeta|_2^2}{2\sigma^2}+\frac{\lambda}{\sigma}\sum_{j=1}^p {|\bX_{:,j}|}_2|\beta_j|\Big\},
\end{equation}
where $\lambda>0$ is a scale-free tuning parameter controlling the trade-off between data fitting and sparsity level.
After a change of variables, this can be cast as a convex program. Hence, it is possible to find the global minimum
relatively efficiently even for large $p$.

A second method for joint estimation of $\bbeta^*$ and $\sigma^*$ by convex programming, the Square-Root Lasso \cite{Antoniadis,Belloni}, estimates
$\bbeta^*$ by $\widehat\bbeta{}^{\text{SqR-L}}$ which solves
\begin{equation}\label{eq:3}
\min_{\bbeta} \Big\{ {\big|\bY-\bfX\bbeta\big|}_2+{\lambda}\sum_{j=1}^p {|\bX_{:,j}|}_2|\beta_j|\Big\}
\end{equation}
and then defines $\widehat\sigma^{\text{SqR-L}}=\frac1{\sqrt{T}}{\big|\bY-\bfX\widehat\bbeta{}^{\text{SqR-L}}\big|}_2$
as an estimator of $\sigma^*$. Both in theory and in practice, these two methods perform quite similarly \cite{Sun}.

%On the one hand, the scaled Lasso has the advantage of being defined via well known statistical protocol:
%penalized (negative) log-likelihood minimization, whereas the Square-root Lasso minimizes
%a cost function whose definition seems dictated only by mathematical convenience.
%On the other hand, the Square-root Lasso requires to solve an SOCP, while the scaled Lasso involves
%a general convex program.

A third method, termed scaled Dantzig selector, was studied by \cite{DalChen} under a more general type of sparsity assumption
(called fused or indirect sparsity). Inspired by these works, we propose a new procedure for joint estimation of the conditional
mean and the conditional variance in the context of heteroscedasticity and group-sparsity.

\section{Definition of the procedure}

Our methodology originates from the penalized log-likelihood minimization.
Assuming errors $\xi_t$ are i.i.d.\ Gaussian $\mathcal N(0,1)$ and setting $\sff(\bx)=\sum_{j=1}^p \phi_j \sff_j(\bx)$,
the penalized log-likelihood used for defining the group-Lasso estimator is (up to summands independent of $(\sff,\sfr)$):
\begin{align}\label{eq:4}
\text{PL}(\sff,\sfr)= &\sum_{t\in\mcT}\!\!\Big\{\!\!-\!\log(\sfr(\bx_t))+\frac12 {\big(\sfr(\bx_t) y_t-\bX_{t,:}\pphi\big)^2}\Big\}\nonumber\\
&+\sum_{k=1}^K\lambda_k\Big|\sum_{j\in G_k}\nolimits \bX_{:,j}\phi_j\Big|_2,
\end{align}
where $\llambda=(\lambda_1,\ldots,\lambda_K)\in\RR^K_+$ is a tuning parameter. A first strategy for
estimating $(\sffstar,\sfrstar)$ is to minimize $\text{PL}(\sff,\sfr)$ with respect to $\pphi\in\RR^p$ and $\sfr\in
\{\sfg:\RR^d\to\RR: \sfg(\bx)\ge 0,\ \text{ for almost all } \bx\in\RR^d\}$. In view of assumption (A2), we can replace
$\sfr$ by $\sum_{\ell=1}^q \alpha_\ell \sfr_\ell$ with an unknown $q$-vector $\aalpha$.

If we introduce the $T\times q$ matrix $\bfR$ having as generic entry $\sfr_\ell(\bx_t)$, (\ref{eq:4}) translates into a convex program with respect to the
$p+q$ dimensional parameter $(\pphi,\aalpha)\in\RR^p\times\RR^q$, in which the cost function is:
\begin{align}\label{eq:5}
\text{PL}(\pphi,\aalpha)= &\sum_{t\in\mcT}\!\Big\{\!\!-\!\log(\bR_{t,:}\aalpha)+\frac12 {\big( y_t\bR_{t,:}\aalpha -\bX_{t,:}\pphi\big)^2}\Big\}\nonumber\\
&+\sum_{k=1}^K{\lambda_k}\big|\bX_{:,G_k}\pphi_{G_k}\big|_2,
\end{align}
and the constraint $\min_t \bR_{t,:}\aalpha\ge 0$ should be imposed to guarantee that the logarithm is well defined.
This is a convex optimization problem, but it does not fit well the framework under which the convergence guarantees
of the state-of-the-art optimization algorithms are established. Indeed, it is usually required that the smooth components
of the cost function have Lipschitz-smooth derivative, which is not the case for (\ref{eq:5}) because of the presence of the
logarithmic terms. One can circumvent this drawback by smoothing these terms\footnote{This will result in introducing new
parameters the tuning of which may increase the difficulty of the problem.}, but we opted for another solution that relies on
an argument introduced in \cite{Candes06} for justifying the Dantzig selector. Let $\Ppi_{G_k}=\bfX_{:,G_k}(\bfX_{:,G_k}^\top\bfX_{:,G_k})^+\bfX_{:,G_k}^\top$
be the orthogonal projector onto the range of $\bfX_{:,G_k}$ in $\RR^\mcT$\!.

\begin{e-definition}\label{def:1}
Let $\llambda\in\RR^K_+$ be a vector of tuning parameters.
We call the Scaled Heteroscedastic Dantzig selector (ScHeDs) the pair $(\widehat\pphi,\widehat\aalpha)$, where
$(\widehat\pphi,\widehat\aalpha,\widehat \bv)$ is a minimizer w.r.t.\ $(\pphi,\aalpha,\bv)\in\RR^p\times\RR^q\times\RR_+^T$ of the cost function
$$
   \sum_{k=1}^K\nolimits \lambda_k\big|\bfX_{:,G_k}\pphi_{G_k}\big|_2
$$
subject to the constraints
\begin{align}
&\Big|\Ppi_{G_k}\big( \text{\rm diag}(\bY)\bfR\aalpha -\bfX\pphi\big)\big|_2\le {\lambda_k},\
\forall k\in [K];\label{eq:9}\\
& \bfR^\top\bv \le \bfR^\top\text{\rm diag}(\bY) (\text{\rm diag}(\bY)\bfR\aalpha-\bfX\pphi); \label{eq:10}\\
&1/v_t\le \bR_{t,:}\aalpha;\ \forall t\in\mcT.\label{eq:11}
\end{align}
\end{e-definition}

Constraints (\ref{eq:9})-(\ref{eq:11}) are obtained as convex relaxations of the first-order conditions
corresponding to minimizing (\ref{eq:5}). In fact, Eq.\ (\ref{eq:9}) is a standard relaxation for
the condition $\mathbf 0\in\partial_\pphi \text{PL}(\pphi,\aalpha)$, whereas constraints (\ref{eq:10}) and
(\ref{eq:11}) are convex relaxations of the equation $\partial_\aalpha \text{PL}(\pphi,\aalpha)=\mathbf 0$.
Further details on this point are provided in the supplementary material. At this stage and before
presenting theoretical guarantees on the statistical performance of the ScHeDs, we state a result
telling us the estimator we introduced is meaningful.

\begin{theorem}\label{th1}
The ScHeDs is always well defined in the sense that the feasible set of the corresponding optimization problem is not empty:
it contains the minimizer of (\ref{eq:5}).
Furthermore, the ScHeDs can be computed by any SOCP solver.
\end{theorem}

The proof is placed in the supplementary material. As we will see later, thanks to this theorem, we
carried out two implementations of the ScHeDs based on an interior point algorithm and an optimal
first-order proximal method.

\section{Comments on the procedure}\label{sec:4}

\paragraph{Tuning parameters}

One apparent drawback of the ScHeDs is the large number of tuning parameters. Fortunately, some theoretical
results provided in the supplementary material suggest to choose $\lambda_k=\lambda_0\sqrt{r_k}$, where $\lambda_0>0$
is a one-dimensional tuning parameter and $r_k=\text{rank}(\bfX_{:,G_k})$. In particular, when all the predictors within
each group are linearly independent, then one may choose $\llambda$ proportional to the vector $(\sqrt{\text{Card}(G_1)},\ldots,\sqrt{\text{Card}(G_K)})$.

\paragraph{Additional constraints}

In many practical situations one can add some additional constraints to the aforementioned optimization problem without
leaving the SOCP framework. For example, if the response $y$ is bounded by some known constant $L_y$, then it is natural
to look for conditional mean and conditional variance bounded respectively by $L_y$ and $L_y^2$. This amounts to introducing
the (linearizable) constraints $|\bX_{t,:}\pphi|\le L_y \bR_{t,:}\aalpha$ and $\bR_{t,:}\aalpha\ge 1/L_y$ for every
$t\in\mcT$.

\paragraph{Bias correction}

It is well known that the Lasso and the Dantzig selector estimate the nonzero coefficients of the regression vector with a bias
toward zero. It was also remarked in \cite{Sun_2010}, that the estimator of the noise level provided by the scaled
Lasso is systematically over-estimating the true noise level. Our experiments showed the same shortcomings for the ScHeDs. To
attenuate these effects, we propose a two-step procedure that applies the ScHeDs with the penalties $\lambda_k=\lambda_0\sqrt{r_k}$
at the first step and discards from $\bfX$ the columns that correspond to vanishing coefficients of $\widehat\pphi$. At the second
step, the ScHeDs is applied with the new matrix  $\bfX$ and with $\llambda=0$.

\paragraph{Gaussian assumption}

Although the proposed algorithm takes its roots from the log-likelihood of the Gaussian regression, it is by no means necessary
that the noise distribution should be Gaussian. In the case of deterministic design $\bx_t$, it is sufficient to assume that the
noise distribution is sub-Gaussian. For random i.i.d.\ design, arguments similar to those of \cite{Belloni,Gautier} can be applied
to show oracle inequalities for even more general noise distributions.

\paragraph{Equivariance}

Given the historical data $(y_{1:T},\bx_{1:T})$ of the response and the covariates, let us denote by
$\widehat y_{T+1}(y_{1:T})= \big[{\sum_{\ell=1}^p \widehat\phi_j\,\sff_j(\bx_{T+1})}\big]/\big[{\sum_{\ell=1}^q \widehat\alpha_\ell\,
\sfr_\ell(\bx_{T+1})}\big]$ the prediction provided by the ScHeDs for a new observation $\bx_{T+1}$. This prediction is equivariant
with respect to scale change in the following sense. If all the response values $y_1,\ldots,y_T$ are multiplied by some constant $c$,
then it can easily be proved that the new prediction can be deduced from the previous one by merely multiplying it by $c$:
$\widehat y_{T+1}(cy_{1:T})=c\widehat y_{T+1}(y_{1:T})$.
% The proof of this property is simple and, therefore, is left to the reader as exercise.

Most papers dealing with group-sparsity \cite{Lounici_2011,Liu_et_al_2010,Huang_Zhang_2010}
use penalties of the form $\sum_k |\DD_k\pphi_{G_k}|_2$ with some diagonal
matrices $\DD_k$. In general, this differs
from the penalty we use since in our case $\DD_k=(\bfX_{:,G_k}^\top\bfX_{:,G_k})^{1/2}$ is not necessarily diagonal.
Our choice has the advantage of being equivariant w.r.t.\ (invertible) linear transformations of predictors within groups.

%More precisely, if $\widehat\pphi$ is the estimator based on the predictors $\bfX_{:,G_1},\ldots,\bfX_{:,G_K}$
%and $\widetilde\pphi$ is the one based on $\bfX_{:,G_1}\AA_1,\ldots,\bfX_{:,G_K}\AA_K$ for some invertible matrices
%$\AA_1,\ldots,\AA_K$, then $\widehat\pphi_{G_k}=\AA_k\widetilde\pphi_{G_k}$ for every $k$. Consequently, the predictions
%based on these two estimators will be the same.
Interestingly, this difference in the penalty definition has an impact on the calibration of the parameters $\lambda_k$:
while the recommended choice is $\lambda_k^2\propto \text{Card}(G_k)$ when diagonal matrices\footnote{Even if the matrices $\DD_k$ are not diagonal
and are chosen exactly as in our case, recent references like \cite{Simon12} suggest to use $\lambda_k^2\propto \text{Card}(G_k)$ without theoretical
support.} $\DD_k$  are used, it is $\lambda_k^2\propto \text{rank}(\bfX_{:,G_k})$ for the ScHeDs. Thus, the penalty
chosen for the ScHeDs is slightly smaller than that of the usual group-Lasso, which also leads to a tighter risk bound.

\begin{table*}
\small
\begin{center}
\begin{tabular}{c|rr|rr|rr||rr|rr|rr}
\toprule
 &\multicolumn{6}{c||}{\textbf{ScHeDs}} & \multicolumn{6}{c}{\textbf{Square-root Lasso}}\\
 \midrule
 &\multicolumn{2}{c|}{$|\widehat\bbeta-\bbeta^*|_2$} & \multicolumn{2}{c|}{$|\widehat\sparsity-\sparsity^*|$} & \multicolumn{2}{c||}{$10|\widehat\sigma-\sigma^*|$} &
 \multicolumn{2}{c|}{$|\widehat\bbeta-\bbeta^*|_2$} & \multicolumn{2}{c|}{$|\widehat\sparsity -\sparsity^*|$} & \multicolumn{2}{c}{$10|\widehat\sigma-\sigma^*|$} \\
 \midrule
$(T,\hphantom{.}p,\hphantom{.}\sparsity^*,\hphantom{.}\sigma^*)$
                 & \scriptsize\texttt{Ave} & \scriptsize\texttt{StD} & \scriptsize\texttt{Ave}
                 & \scriptsize\texttt{StD} & \scriptsize\texttt{Ave} & \scriptsize\texttt{StD}
                 & \scriptsize\texttt{Ave} & \scriptsize\texttt{StD} & \scriptsize\texttt{Ave}
                 & \scriptsize\texttt{StD} & \scriptsize\texttt{Ave} & \scriptsize\texttt{StD} \\
\midrule
$(100, \hphantom{0}100, 2, 0.5)$ &     .06 &     .03 &     .00 &     .00 &     .29 &     .21 &     .08 &     .05 &     .19 &     .42 &     .31 &     .23\\
$(100, \hphantom{0}100, 5, 0.5)$ &     .11 &     .06 &     .00 &     .00 &     .29 &     .31 &     .12 &     .05 &     .16 &     .41 &     .30 &     .24\\
$(100, \hphantom{0}100, 2, 1.0)$ &     .13 &     .07 &     .02 &     .14 &     .53 &     .40 &     .16 &     .11 &     .19 &     .44 &     .56 &     .42\\
$(100, \hphantom{0}100, 5, 1.0)$ &     .28 &     .24 &     .08 &     .32 &     .76 &     .78 &     .25 &     .13 &     .19 &     .44 &     .66 &     .57\\
$(200, \hphantom{0}100, 5, 0.5)$ &     .08 &     .03 &     .00 &     .00 &     .20 &     .16 &     .09 &     .03 &     .20 &     .46 &     .22 &     .16\\
$(200, \hphantom{0}100, 5, 1.0)$ &     .15 &     .05 &     .01 &     .09 &     .40 &     .30 &     .17 &     .07 &     .20 &     .44 &     .42 &     .31\\
$(200, \hphantom{0}500, 8, 0.5)$ &     .10 &     .03 &     .00 &     .04 &     .23 &     .16 &     .11 &     .03 &     .17 &     .40 &     .24 &     .17\\
$(200, \hphantom{0}500, 8, 1.0)$ &     .21 &     .13 &     .02 &     .17 &     .50 &     .58 &     .22 &     .08 &     .19 &     .43 &     .46 &     .38\\
$(200, 1000, 5,1.0)$ &     .15 &     .05 &     .01 &     .08 &     .40 &     .31 &     .17 &     .07 &     .17 &     .40 &     .42 &     .33\\
\bottomrule
\end{tabular}
\caption{\small{Performance of the (bias corrected) ScHeDs compared with the (bias corrected) Square-root Lasso on a synthetic dataset. The average values and
the standard deviations of the quantities $|\widehat\bbeta-\bbeta^*|_2$, $|\widehat\sparsity-\sparsity^*|$ and $10|\widehat\sigma-\sigma^*|$
over 500 trials are reported. They represent respectively the accuracy in estimating the regression vector, the number of
relevant covariates and the level of noise.}\vspace{-20pt}}
\label{tab:1}
\end{center}
\end{table*}

\section{Risk bounds}

We present a finite sample risk bound showing that, under some assumptions, the risk of our procedure is of the same order of
magnitude as the risk of a procedure based on the complete knowledge of the noise-level.

Recall that the model introduced in the foregoing sections can be rewritten
in its matrix form
\begin{equation}
\label{modelematrice}
 \text{\rm diag}(\bY)\bfR\aalpha^*= \bfX\pphi^*+\xxi,
\end{equation}
with $\xi_1,\ldots,\xi_T$ i.i.d.\ zero mean random variables.
To state the theoretical results providing guarantees on the accuracy of the ScHeDs
estimator $(\widehat{\pphi},\widehat{\aalpha})$, we need some notation and assumptions.

For $\pphi^* \in\mathbb{R}^p$, we define the set of relevant groups $\mathcal K^*$ and the sparsity index $\sparsity^*$ by
\begin{align}
\label{K0}
\mathcal{K}^*&= \left\{k: \big| \pphi_{G_k}^*\big|_1\neq 0\right\},\  \sparsity^*=\sum_{k\in{\mcK^*}}\nolimits r_k,
\end{align}
Note that these quantities depend on $\pphi^*$. To establish tight risk bounds, we need the following assumption on the
Gram matrix $\bfX^\top\bfX$,  termed Group-Restricted Eigenvalues (GRE).

\textbf{Assumption GRE$(N,\kappa)$:} For every $\mcK\subset[p]$ of cardinality not larger than $N$ and for every
$\ddelta\in\mathbb{R}^p$ satisfying
\begin{equation}\label{ineq1}
  \sum_{\mathcal{K}^c}\nolimits  \lambda_k\big|\bfX_{:,G_k}\ddelta_{G_k}\big|_2\leq
  \sum_{\mathcal{K}}\nolimits \lambda_k\big|\bfX_{:,G_k}\ddelta_{G_k}\big|_2,
\end{equation}
it holds that
%\begin{equation}\label{assump1}
$  \big|\bfX\ddelta\big|_2^2\geq\kappa^2\sum_{k\in\mathcal{K}}\big|\bfX_{:,G_k}\ddelta_{G_k}\big|^2_2.$
%\end{equation}

We also set
\begin{align}
\label{constante}
 C_1&=\max_{\ell=1,\ldots, q}{ \frac1T \sum_{t\in\mcT}\nolimits  \frac{r_{t\ell}^2(\bX_{t,:}\pphi^*)^2}{(\bR_{t,:}\aalpha^*)^2}\ },
\\
%\end{align}
%and
%\begin{align}
 \label{constante2}
C_2&=\max_{\ell=1,\ldots, q}{ \frac1T \sum_{t\in\mcT}\nolimits  \frac{r_{t\ell}^2}{(\bR_{t,:}\aalpha^*)^2}\ },\\
\label{constante3}
C_3&=\min_{\ell=1,\ldots, q}\frac1T \sum_{t\in\mcT}\nolimits  \frac{r_{t\ell}}{(\bR_{t,:}\aalpha^*)},
\end{align}
and define $C_4=(\sqrt{C_2}+\sqrt{2C_1})/C_3$.

To establish nonasymptotic risk bounds in the heteroscedastic regression model with sparsity assumption, we first tried
to adapt the standard techniques \cite{Candes06,BRT2009} used in the case of known noise-level. The result, presented in
Theorem~\ref{th2} below, is not satisfactory, since it provides a risk bound for estimating $\pphi^*$ that involves the risk
of estimating $\aalpha^*$. Nevertheless, we opted for stating this result since it provides guidance for choosing the
parameters $\lambda_k$ and also because it constitutes an important ingredient of the proof of our main result stated in
Theorem~\ref{th3} below.

\begin{theorem}
\label{th2}
Consider model \eqref{modelematrice} with deterministic matrices $\bfX$ and $\bfR$. Assume that the distribution of $\xxi$ is Gaussian
with zero mean and an identity covariance matrix and that Assumption~GRE$(K^*,\kappa)$ is fulfilled with $K^*=\text{\rm Card}(\mcK^*)$.
Let $\varepsilon\in(0,1)$ be a tolerance level and set
$$
\lambda_k= 2\big(r_k+2\sqrt{r_k \log(K/\varepsilon)}+2\log(K/\varepsilon)\big)^{1/2}.
$$
If  $T\ge 8C_4\log(\frac{2q}{\varepsilon})$ then,  with probability at least $1-2\varepsilon$,
\begin{align}
\big| \bfX(\widehat{\pphi}-\pphi^*) \big| _2&\leq C_4\sqrt{{(8/T)\log(2q/\varepsilon)}} (2| \bfX\pphi^* |_2+|\xxi|_2)\nonumber\\
&\qquad + \frac{8}{\kappa}\sqrt{2 \sparsity^*+3K^* \log(K/\varepsilon)}\nonumber\\
&\qquad + |\diag(\bY)\bfR(\widehat{\aalpha}-\aalpha^*) |_2.\label{bound:1}
\end{align}
\end{theorem}

In order to gain understanding on the theoretical limits delineated by the previous theorem, let
%us consider the situation of bounded functions $\sfb^*$, $\sfsstar$ and $1/\sfsstar$. This assumption implies that both $|\bfX\pphi^*|_\infty$ and
%$1/\min_i|\bfR_i\aalpha^*|$ are bounded by some constant $C_*$. Let
us give more details on the order of magnitude of the three terms appearing in (\ref{bound:1}). First, one should keep in mind that the correct
normalization of the error consists in dividing $\big| \bfX(\widehat{\pphi}-\pphi^*) \big| _2$ by $\sqrt{T}$. Assuming that the function $\sfb^*$
is bounded and using standard tail bounds on the $\chi^2_T$ distribution, we can see that the first term in the right-hand side of (\ref{bound:1})
is negligible w.r.t.\ the second one. Thus if we ignore for a moment the third term, Theorem~\ref{th2} tells us that the normalized squared error
$T^{-1}\big|\bfX(\widehat{\pphi}-\pphi^*)\big|_2^2$ of estimating $\pphi^*$ by $\widehat\pphi$ is of the order of $\sparsity^*/T$, up to logarithmic terms. This is the (optimal)
fast rate of estimating an $\sparsity^*$-sparse signal with $T$ observations in linear regression.

To complete the theoretical analysis, we need a bound on the error of estimating the parameter $\aalpha^*$. This is done in the following theorem.
%From now on, $C$ is an absolute constant that may vary
%from line to line. We would like to remind the reader that the optimal rate of converge for prediction error is of order
%$\sqrt{ \big|\pphi^*\big|_0\log(pq)/T}$, where$ \big|\pphi^*\big|_0$ is the
%number of non-zero elements in $\pphi^*$.
%The following theorem provides an upper bound  for the prediction error of the estimator $(\widehat{\pphi},\widehat{\aalpha})$.

\begin{theorem}\label{th3}
Let all the conditions of Theorem~\ref{th2} be fulfilled. Let $q$ and $T$ be two integers such that $1\le q\le T$ and let $\varepsilon\in(0,1/5)$.
Assume that for some constant ${\hat D_1}\ge 1$ the inequality $\max_{t\in\mcT}\frac{\bR_{t,:}\hat\aalpha}{\bR_{t,:}\aalpha^*}\le {\hat D_1}$ holds true and
denote $D_{T,\varepsilon}={\hat D_1}(2|\bfX\pphi^*|_\infty^2+5{\log(2T/\varepsilon)})$. Then, on an event of probability at least
$1-5\varepsilon$, the following inequality is true:
\begin{align}
\label{bound:2}
\big| \bfX(\widehat{\pphi}-\pphi^*) \big| _2
& \leq 4 (C_4+1)D_{T,\varepsilon}^{3/2}\sqrt{{2q\log(2q/\varepsilon)}}\nonumber\\
&+ \frac{8D_{T,\varepsilon}}{\kappa}\sqrt{2 \sparsity^*+3K^* \log(K/\varepsilon)}.
\end{align}
Furthermore, on the same event,
\begin{align}
\label{bound:3}
\frac{\big|\bfR(\aalpha^*-\hat\aalpha)\big|_2}{{\hat D_1}^{1/2} |\bfR\aalpha^*|_\infty}
    &\le 4 (C_4+2)D_{T,\varepsilon}^{3/2}\sqrt{{2q\log(2q/\varepsilon)}}\nonumber\\
    &+ \frac{8D_{T,\varepsilon}}{\kappa}\sqrt{2 \sparsity^*+3K^* \log(K/\varepsilon)}.
\end{align}
\end{theorem}

The first important feature of this result is that it provides fast rates of convergence
for the ScHeDs estimator. This compares favorably with the analogous result in \cite{Kolar12},
where asymptotic bounds are presented under the stringent condition that the local minimum
to which the procedure converges coincides with the global one. The joint convexity in $\pphi$
and $\aalpha$ of our minimization problem allows us to avoid such an assumption without
any loss in the quality of prediction.

One potential weakness of the risk bounds of Theorem \ref{th3} is the
presence of the quantity $\widehat D_1$, which controls, roughly speaking, the
$\ell_\infty$ norm of the vector $\bfR\hat\aalpha$. One way to circumvent this drawback is
to add the constraint $\max_t \bR_{t,:}\aalpha \le \mu^*$ to those presented in
(\ref{eq:9})-(\ref{eq:11}), for some tuning parameter $\mu^*$. In this case, the optimization
problem remains an SOCP and in all the previous results one can replace the random term $\hat D_1$ by
$\mu^*/\mu_*$, where $\mu_*$ is a lower bound on the elements of the vector $\bfR\aalpha^*$. This
being said, we hope that with more sophisticated arguments one can deduce the boundedness
of $\hat D_1$ by some deterministic constant without adding new constraints to the ScHeDs.

One may also wonder how restrictive the assumptions (\ref{constante})-(\ref{constante3}) are and
in which kind of contexts they are expected to be satisfied. At a heuristic level, one may remark
that the expressions in (\ref{constante})-(\ref{constante3}) are all empirical means: for instance,
$(1/T)\sum_t r_{t\ell}^2/(\bR_{t,:}\aalpha^*)^2 =(1/T)\sum_t \sfr_\ell(\bx_t)^2/\sfrstar(\bx_t)^2$.
Assuming that the time series $\{\bx_t\}$ is stationary or periodic, these empirical means will
converge to some expectations. Therefore, under these types of assumptions, (\ref{constante})-(\ref{constante3})
are boundedness assumptions on some integral functionals  of $\sfrstar$, $\sffstar$ and $\sfr_\ell$'s.
In particular, if $\sfr_\ell$'s are bounded and bounded away from 0, $\sffstar$ is bounded and $\sfrstar$
is bounded away from zero, then the finiteness of the constant $C_4$ is straightforward.

To close this section, let us emphasize that the GRE condition is sufficient for getting fast rates
for the performance of the ScHeDs measured in prediction loss, but is by no means necessary for the
consistency. In other terms, even if the GRE condition fails, the ScHeDs still provides provably accurate
estimates that converge at a slower rate. This slow rate is, roughly speaking, of the order
$[T^{-1}(\sparsity^*+K^* \log K)]^{1/4}$ instead of $[T^{-1}(\sparsity^*+K^* \log K)]^{1/2}$.

\section{Experiments}

To assess the estimation accuracy of our method and to compare it with the state-of-the-art alternatives, we performed
an experiment on a synthetic dataset. Then, the prediction ability of the procedure is evaluated on a
real-world dataset containing the temperatures in Paris over several years.

\subsection{Implementation}

\begin{figure}
\includegraphics[width=0.5\textwidth]{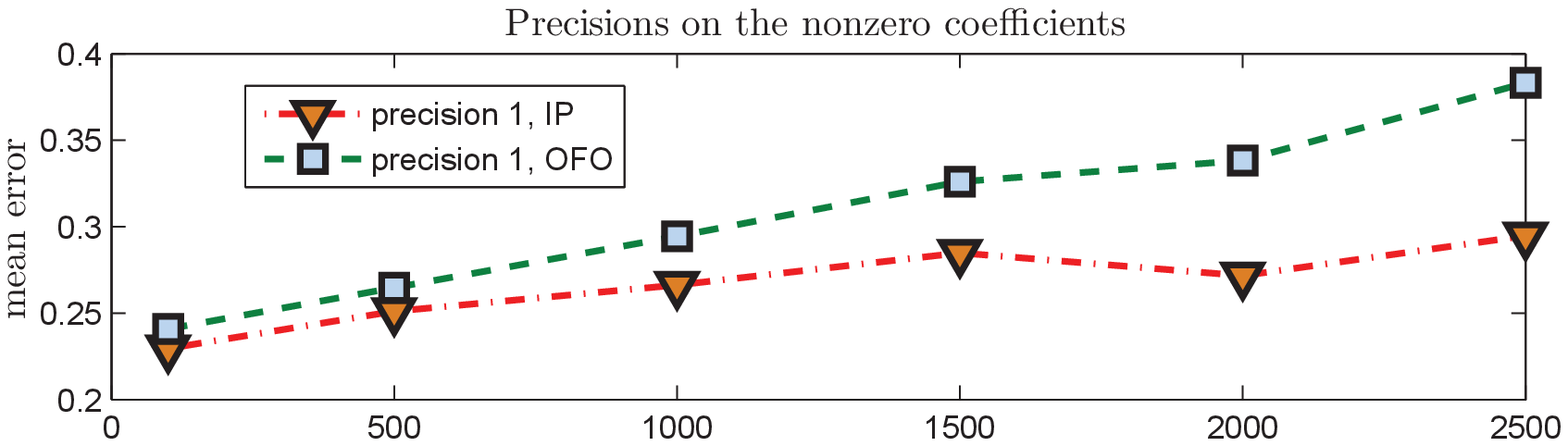}\\
\includegraphics[width=0.5\textwidth]{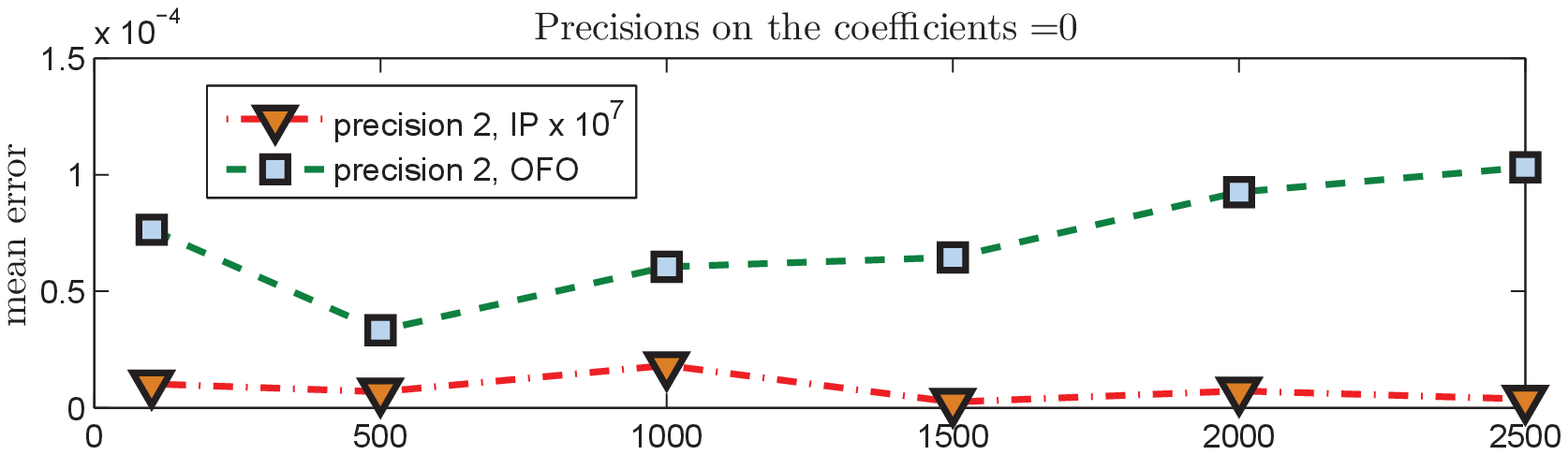}\\
\includegraphics[width=0.5\textwidth]{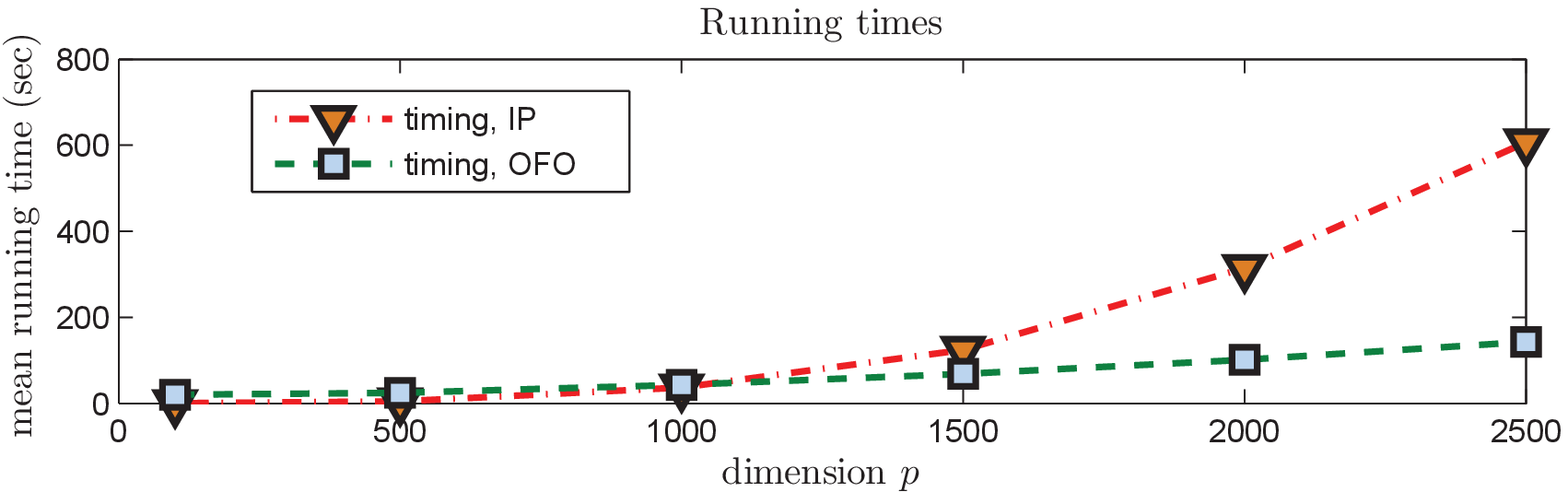}
\vspace{-20pt}
\caption{\footnotesize  Comparing implementations of the ScHeDs: interior point (IP) \textit{vs.}\ optimal first-order (OFO) method.
 We used the experiment described in Section~\ref{ssec:exp1} with $T=200$, $\sparsity^*=3$, $\sigma=0.5$.
 Top: square-root of the MSE on the nonzero coefficients of $\bbeta^*$.
 Middle:  square-root of the MSE on the zero coefficients of $\bbeta^*$.
 Bottom: running times.  \vspace{-15pt}}
\label{fig:2}
\end{figure}

To effectively compute the ScHeDs estimator we rely on Theorem~\ref{th1} that reduces the computation
to solving a second-order cone program. To this end, we implemented a primal-dual interior point method
using the SeDuMi package \cite{SeDuMi} of Matlab as well as several optimal first-order methods
\cite{Nesterov83,Aus_Teb06,Beck_Teb09} using the TFOCS \cite{TFOCS}. We intend to make our code
publicly available if the paper is accepted. Each of these implementations has its strengths and limitations.
The interior point method provides a highly accurate solution for moderately large datasets (Fig.~\ref{fig:2}, top), but this accuracy
is achieved at the expense of increased computational complexity (Fig.~\ref{fig:2}, bottom). Although less accurate, optimal first-order
methods have cheaper iterations and can deal with very large scale datasets (see Table~\ref{tab:2}).
All the experiments were conducted on an  Intel(R) Xeon(R) CPU \PVerb{@}2.80GHz.

\subsection{Synthetic data}\label{ssec:exp1}

In order to be able to compare our approach to other state-of-the-art algorithms, we place ourselves
in a setting of homoscedastic noise with known ground truth. We randomly generate a matrix $\bfX\in\RR^{T\times p}$ with
i.i.d.\ standard Gaussian entries and a standard Gaussian noise vector $\xxi\in\RR^{T}$ independent of $\bfX$. The
noise variance is defined by $\sigma_t\equiv\sigma^*$ with varying values $\sigma^*>0$. We set
$\bbeta^0=[\mathbf 1_{\sparsity^*},\ \mathbf 0_{p-\sparsity^*}]^\top$ and define $\pphi^*=\bbeta^*/\sigma^*$, where $\bbeta^*$ is
obtained by randomly permuting the entries of $\bbeta^0$.  Finally, we set $\bY=\sigma^*(\bfX\pphi^*+\xxi)$.

Nine different settings depending on the values of $(T,p,\sparsity^*,\sigma^*)$ are considered. In each setting the experiment
is repeated 500 times; the average errors of estimation of $\bbeta^*$, $\sparsity^*$ and $\sigma^*$ for our procedure and
for the Square-root Lasso are reported in Table~\ref{tab:1} along with the standard deviations. For both procedures,
the universal choice of tuning parameter $\lambda=\sqrt{2\log(p)}$ is used (after properly normalizing the columns of $\bfX$)
and a second step consisting in bias correction is applied (\lcf \cite{Sun} and the discussion in Section ~\ref{sec:4} on
bias correction). Here, we did not use any group structure so the penalty is merely proportional to the $\ell_1$-norm of
$\bbeta$. One can observe that the ScHeDs is competitive with the Square-root Lasso, especially for performing variable
selection. Indeed, in all considered settings the ScHeDs outperforms the Square-root Lasso in estimating $\sparsity^*$.

\begin{table}
 \small
 \begin{center}
 \begin{tabular}{l|r|r|r|r|r}
 \toprule
 $p$               &  200    & 400     & 600   & 800    & 1000\\
 \midrule
 IP (sec/iter)     &  0.14   & 0.70    & 2.15  & 4.68   & 9.46\\
 OFO (100*sec/iter) &  0.91   & 1.07    & 1.33  & 1.64   & 1.91\\
 \bottomrule
 \end{tabular}
 \caption{\footnotesize  Comparing implementations of the ScHeDs: interior point (IP) \textit{vs.}\ optimal first-order (OFO) method.
 We report the time per iteration (in seconds) for varying $p$ in the experiment described in
 Section~\ref{ssec:exp1} with $T=200$, $\sparsity^*=2$, $\sigma=0.1$. Note that the iterations of the OFO are
 very cheap and their complexity increases linearly in $p$. \vspace{-15pt}}
 \label{tab:2}
 \end{center}
 \end{table}

\subsection{Application to the prediction of the temperature in Paris}
\begin{figure*}
\begin{center}
\includegraphics[width=390pt]{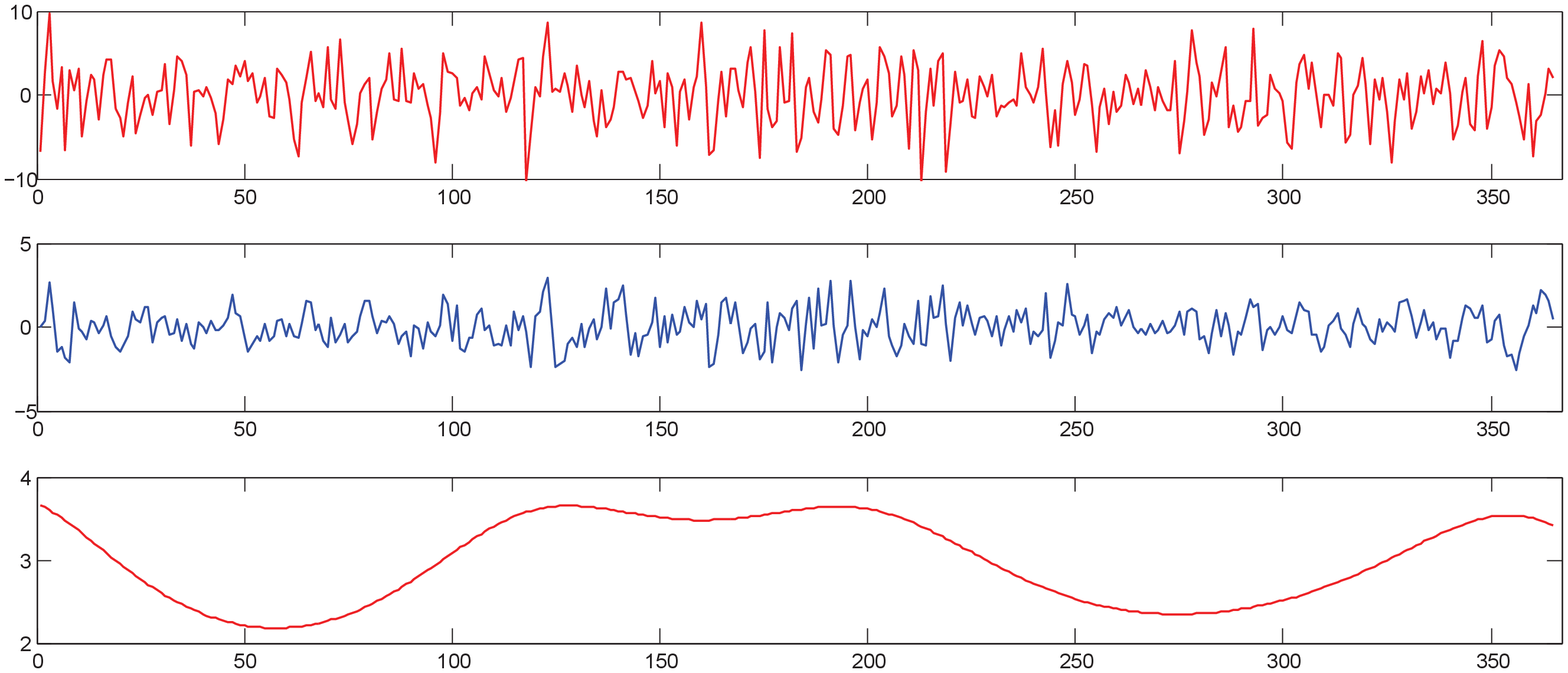}
\vspace{-10pt}
\caption{\footnotesize  Top row: increments of temperatures (in Fahrenheit) from one day to the next observed in Paris in 2008.
Middle row: predictions provided by our ScHeDs procedure; we observe that the sign is often predicted correctly. Bottom row:
estimated noise level.\vspace{-15pt}}
\label{fig:1}
\end{center}
\end{figure*}

For experimental validation on a real-world dataset, we have used data
on the daily temperature in Paris from
2003 to 2008. It was produced by the National Climatic Data Center (NCDC), (Asheville, NC, USA) and
is publicly available at \url{ftp://ftp.ncdc.noaa.gov/pub/data/gsod/}. Performing good predictions
for these data is a challenging task since, as shown in Fig.~\ref{fig:1}, the observations look like white
noise. The dataset contains the daily average temperatures, as well as some other measurements like
wind speed, maximal and minimal temperatures, \textit{etc}.

We selected as response variable $y_t$ the difference of temperatures between two successive days. The goal was to
predict the temperature of the next day%---or, equivalently, the increment of the temperature---
based on historical data.
We selected as covariates $\bx_t$ the time $t$, the increments of temperature over past 7 days, the maximal intraday variation
of the temperature over past 7 days and the wind speed of the day before. Including the intercept, this resulted in a 17
dimensional vector $\bx_t$. Based on it, we created $136$ groups of functions $\sff$, each group containing $16$ elements.
Thus, the dimension of $\pphi^*$ was $136\times 16 = 2176$. We chose $q=11$ with functions $\sfr_\ell$ depending on time $t$
only. The precise definitions of $\sff_j$ and $\sfr_\ell$ are presented below.

%\paragraph{Definition of $\bfX$}

To specify $\bfX$, we need to define the functions $\sff_j$ generating its columns.
We denote by $\bu_t$ the subvector of $\bx_t$ obtained by removing the time $t$. Thus, $\bu_t$ is a 16-dimensional vector.
%containing the daily
%increments of temperatures over past 7 days, the intraday variations of temperatures over
%7 days, the wind speed and the constant factor.
Using this vector $\bu_t\in\RR^{16}$, we define all the second-order monomes:
$\chi_{i,i'}(\bu_t)=u_t^{(i)}u_t^{(i')}$ with $i \le i'$. We look for fitting the unknown function
$\sffstar$ by a second-order polynomial in $\bu_t$ with coefficients varying in time.
To this end, we set $\psi_1(t)= 1$, $\psi_\ell(t) = t^{1/(\ell-1)}$, for $\ell=2,3,4$ and
\begin{align*}
\psi_\ell(t) &= \cos(2\pi(\ell-4) t/365);\qquad \ell=5,\ldots,10;\\
\psi_\ell(t) &= \sin(2\pi(\ell-10) t/365);\qquad \ell=11,\ldots,16.
\end{align*}
Once these functions $\chi_{i,i'}$ and $\psi_\ell$ defined, we denote by $\sff_j$
the functions of the form $\psi_\ell(t)\chi_{i,i'}(\bu_t)$. In other terms, we
compute the tensor product of these two sets of functions, which leads to a set
of functions $\{\sff_j\}$ of cardinality $16\times 16\times 17/2 = 2176$. These
functions are split into 136 groups of 16 functions, each group defined by
$G_{i,i'}=\{\psi_\ell(t)\times\chi_{i,i'}(\bu_t):\ell=1,\ldots,16\}$.

%\paragraph{Definition of $\bfR$}

We defined $\bfR$ as a $T\times 11$ matrix, each of its eleven columns was obtained
by applying some function $\sfr_\ell$ to the covariate $\bx_t$ for $t=1,\ldots,T$. The
functions $\sfr_\ell$ were chosen as follows: $\sfr_1(\bx_t) = 1$, $\sfr_2(\bx_t) = t$,
$\sfr_3(\bx_t) = 1/(t+2\times 365)^{\frac12}$ and
\begin{align*}
\sfr_\ell(\bx_t) &= 1+\cos(2\pi(\ell-3) t/365);\qquad \ell=4,\ldots,7;\\
\sfr_\ell(\bx_t) &= 1+\cos(2\pi(\ell-7) t/365);\qquad \ell=8,\ldots,11.
\end{align*}
Note that these definitions of $\bfX$ and $\bfR$ are somewhat arbitrary. Presumably, better
results in terms of prediction would be achieved by combining this purely statistical approach
with some expert advice.

We used the temperatures from 2003 to 2007 for training (2172 values) and those of 2008 (366 values) for testing. Applying
our procedure allowed us to reduce the dimensionality of $\pphi$ from $2176$ to $26$. The result of the prediction for the
increments of temperatures in 2008 is depicted in Fig.~\ref{fig:1}. The most important point is that in 62\% of the cases the
sign of the increments is predicted correctly. It is also interesting to look at the estimated variance: it suggests that
the oscillation of the temperature during the period between  May and July is significantly higher than in March, September
and October.  %A last observation concerns the distribution of the residuals for the test dataset.
Interestingly, when we apply a
Kolmogorov-Smirnov test to the residuals $y_t\bR_{t,:}\widehat\aalpha-\bX_{t,:}\widehat\pphi$ for $t$ belonging to the testing
set, the null hypothesis of Gaussianity is not rejected and the $p$ value is $0.72$.

%\section{Sketch of the proof}
%
%
%{\color{blue} One half page describing main steps of the proof.
%
%The complete proof will go to the supplementary material.}

\section{Conclusion and outlook}

We have introduced a new procedure, the ScHeDs, that allows us to simultaneously estimate the
conditional mean and the conditional variance functions in the model of regression
with heteroscedastic noise. The ScHeDs relies on minimizing a group-sparsity
promoting norm under some constraints corresponding to suitably relaxed first-order
conditions for maximum penalized likelihood estimation. We have proposed several implementations
of the ScHeDs based on various algorithms of second-order cone programming. We have tested our
procedure on synthetic and real world datasets and have observed that it is competitive with
the state-of-the-art algorithms, while being applicable in a much more general framework.
Theoretical guarantees for this procedure have also been proved.

In a future work, we intend to generalize this approach to the case where the
inverse of the conditional standard deviation belongs to a reproducing kernel Hilbert space, or admits
a sparse linear representation in a large, possibly over-complete, dictionary.
The extension of our methodology to the case of nonoverlapping groups
\cite{Obozinski_2011,Mairal_2011} and the substitution of the $\ell_1/\ell_2$-norm
penalty by more general $\ell_1/\ell_q$-norms in our framework are %other
challenging avenues for future research.

% \appendix

% \section{}
% \label{}

% The Acknowledgements are an un-numbered section
%\section*{Acknowledgements}
% Acknowledgements text here

%\newpage
{\small
\setlength{\baselineskip}{0.98\baselineskip}

\bibliography{ref_all}
\bibliographystyle{icml2013}
}

\newpage

\onecolumn

\section{Supplement to the paper: Learning Heteroscedastic Models by Convex Programming under Group Sparsity}

%{\color{blue}
%
%%%%%%%%%%%%%%%%%%%%%%%%%%%%%%%%
%\subsection{Additional details on the experiment on the temperature dataset}
%
%The experimental settings described in the main paper complete, except that
%the precise definition of matrices $\bfX$ and $\bfR$ is not provided. The aim of
%this section is to fill this gap.
%
%}

%%%%%%%%%%%%%%%%%%%%%%%%%%%%%%%
\subsection{Proof of Theorem~\ref{th1}}

The fact that the feasible set is not empty follows from the fact that it contains the minimizers of (\ref{eq:5}). This immediately
follows from the first-order conditions and their relaxations. Indeed, for a minimizer $(\pphi^\circ,\aalpha^\circ)$ of (\ref{eq:5}),
the first-order conditions take the following form: there exists $\nnu^\circ\in \RR^T_+$
such that for all $k\in[K]$ and $\ell\in[q]$,
\begin{align}
\label{eq:6}
\frac{\partial}{\partial \phi_{G_k}} \text{PL}(\pphi^\circ,\aalpha^\circ)
&= -\bfX_{:,G_k}^\top{\big( \text{diag}(\bY)\bfR\aalpha^\circ -\bfX\pphi^\circ\big)}+{\lambda_k}\bfX_{:,G_k}^\top
\frac{\bfX_{:,G_k}\pphi_{G_k}^\circ}{{\big|\bfX_{:,G_k}\pphi_{G_k}^\circ\big|}_2}=0,\\
\label{eq:7}
\frac{\partial}{\partial \alpha^\circ_\ell} \text{PL}(\pphi^\circ,\aalpha^\circ)
&= -\sum_{t\in\mcT}\nolimits \frac{r_{t\ell}}{\bR_{t,:}\aalpha^\circ}+\sum_{t\in\mcT} \nolimits{\big( y_t\bR_{t,:}\aalpha^\circ -\bX_{t,:}\pphi^\circ\big)y_tr_{t\ell}}
-(\nnu^\circ)^\top\bR_{:,\ell}=0,
\end{align}
and $\nu^\circ_t\bR_{t,:}\aalpha^\circ=0$ for every $t$.
It should be emphasized that relation (\ref{eq:6}) holds true only in the case where the solution satisfies $\min_k|\bX_{:,G_k}\pphi^\circ_{:,G_k}|_2\not=0$,
otherwise one has to replace it by the condition stating that the null vector belongs to the subdifferential. Since this does not alter the proof, we prefer to
proceed as if everything was differentiable.

%If we consider equations (\ref{eq:6}) and (\ref{eq:7}) as a system of equations w.r.t.\ the unknowns $\pphi^\circ$ and $\aalpha^\circ$, it is
%difficult to solve because of its inherent nonlinearity. The key idea leading to our procedure is to relax condition (\ref{eq:6})
%by observing that
On the one hand, $(\pphi^\circ,\aalpha^\circ)$ satisfies (\ref{eq:6}) if and only if $\Ppi_{G_k}(\text{diag}(\bY)\bfR\aalpha^\circ-\bfX\pphi^\circ)=\lambda_k \bfX_{:,G_k}\pphi^\circ_{G_k}/|\bfX_{:,G_k}\pphi^\circ_{G_k}|_2$ with $\Ppi_{G_k}=\bfX_{:,G_k}(\bfX_{:,G_k}^\top\bfX_{:,G_k})^+\bfX_{:,G_k}^\top$
being the orthogonal projector onto the range of $\bfX_{:,G_k}$ in $\RR^T$.
Taking the norm of both sides in the last equation, we get
$%\begin{equation}\label{eq:8}
 \big|\Ppi_{G_k}(\text{diag}(\bY)\bfR\aalpha^\circ-\bfX\pphi^\circ)\big|_2\le \lambda_k.
$ %\end{equation}
This tells us that $(\pphi^\circ,\aalpha^\circ)$ satisfy (\ref{eq:9}). On the other hand, since the minimum of (\ref{eq:5}) is finite,
one easily checks that $\bR_{t,:}\aalpha^\circ\not =0$ and, therefore, $\nnu^\circ =0$. Replacing in (\ref{eq:7}) $\nnu^\circ$ by zero and setting
$v_t^\circ=1/\bR_{t,:}\aalpha^\circ$ we get that $(\pphi^\circ,\aalpha^\circ,\bv^\circ)$ satisfies (\ref{eq:10}), (\ref{eq:11}). This
proves that the set of feasible solutions of the optimization problem defined in the ScHeDs is not empty.
%Finally, to get the relaxed problem  close to the original one, we look for the solution in the feasible set minimizing the group-sparsity promoting norm
%$\sum_{k=1}^K \lambda_k\big|\bfX_{:,G_k}\pphi_{G_k}\big|_2$.  These considerations lead to the algorithm ScHeDs.

Let us show that one can compute the ScHeDs $(\widehat\pphi,\widehat\aalpha)$ by solving an SOCP. More precisely, we show that if
$(\widehat\pphi,\widehat\aalpha,\widehat\bu,\widehat\bv)\in \RR^p\times\RR^q\times\RR^K\times\RR^T$ is a solution to
the following problem of second-order cone programming:
    \begin{align}\displaystyle
        \qquad \qquad \min&\quad \sum_{k=1}^K\nolimits \lambda_k u_k& &\\
        \text{subject to }&  \text{ (\ref{eq:9}) and }&  & \nonumber\\
        &\big|\bfX_{:,G_k}\pphi_{G_k}\big|_2\le u_k,\label{eq:11c}& \forall k\in[K],&\\
        & \bfR^\top\bv \le \bfR^\top\text{\rm diag}(\bY) (\text{\rm diag}(\bY)\bfR\aalpha-\bfX\pphi); \quad &
        &\label{eq:12}\\
        &\big|\big[v_t;\bR_{t,:}\aalpha;\sqrt{2}\hspace{2pt}\big]\big|_2\le v_t+\bR_{t,:}\aalpha;\qquad\qquad\qquad &\forall t\in\mcT,&
    \end{align}
then $(\widehat\pphi,\widehat\aalpha,\widehat\bv)$ is a solution to the optimization problem stated in Definition~\ref{def:1}. This claim readily follows from the fact
that the constraint $\big|\big[v_t;\bR_{t,:}\aalpha;\sqrt{2}\hspace{2pt}\big]\big|_2\le v_t+\bR_{t,:}\aalpha$ can be equivalently written as
$v_t(\bR_{t,:}\aalpha)\ge 1$ and $v_t+\bR_{t,:}\aalpha\ge 0$ for every $t$. This yields $v_t\ge 0$ and $\bR_{t,:}\aalpha\ge 1/v_t$ for every $t$.
Furthermore, it is clear that if $(\widehat\pphi,\widehat\aalpha,\widehat\bu,\widehat\bv)$ is a solution to the aforementioned optimization problem,
then all the inequalities in (\ref{eq:11c}) are indeed equalities. This completes the proof.

%%%%%%%%%%%%%%%%%%%
\subsection{Proof of  Theorem~\ref{th2}}
%%%%%%%%%%%%%%%%%%%

 To prove Theorem~\ref{th2},  we first introduce a feasible pair $(\tilde{\pphi},\tilde{\aalpha})$, in the sense formulated in Lemma~\ref{lm1}.
\begin{lemma}
\label{lm1}
Consider the model \eqref{modelematrice}. Let  $z=1+2C_4\sqrt{\frac{2\log(2q/\varepsilon)}{T}}$
with some $\varepsilon>0$ and assume that $z\le 2$.  Then with probability at least $1-2\varepsilon$,
the triplet $(\tilde{\pphi},\tilde{\aalpha},\tilde\bv)=\big(z\pphi^*,z\aalpha^*,(\frac1{z\bR_{t,:}\aalpha^*})_{t=1,\ldots,T}\big)$
satisfies constraints \eqref{eq:9}, \eqref{eq:10} and \eqref{eq:11}.
Moreover,  the group-sparsity pattern $ \big\{k: \big|\tilde{ \pphi}_{G_k}\big| _1\neq 0\big\}$ of $\tilde\pphi$ coincides with
that of $\pphi^*$, that is with ${\mcK^*}$.
\end{lemma}

 The proof of this lemma can be found in Section~\ref{prooflm1}.\\

 Set  $\DDelta= \widehat{\pphi}-\tilde{\pphi}$. On an event of probability at least $1-2\varepsilon$, $(\tilde\pphi,\tilde\aalpha)$ is a feasible
 solution of the optimization problem of the ScHeDs whereas $( \widehat{\pphi},\widehat{\aalpha})$ is an optimal solution, therefore
\begin{align}
\label{deux}
\sum_{k=1}^K\lambda_k\big|\bfX_{:,G_k}\DDelta_{G_k}\big|_2&\leq\sum_{k=1}^K\lambda_k\big|\bfX_{:,G_k}\DDelta_{G_k}\big|_2+\sum_{k=1}^K\lambda_k\big|\bfX_{:,G_k}\tilde{\pphi}_{G_k}\big|_2-\sum_{k=1}^K\lambda_k\big|\bfX_{:,G_k}\widehat{\pphi}_{G_k}\big|_2\nonumber\\
&=\sum_{k\in{\mcK^*}}\lambda_k\big|\bfX_{:,G_k}\DDelta_{G_k}\big|_2+\sum_{k\in{\mcK^*}}\lambda_k( \big|\bfX_{:,G_k}\tilde{\pphi}_{G_k}\big|_2 - \big|\bfX_{:,G_k}\widehat{\pphi}_{G_k}\big|_2  )\nonumber\\
&\leq2\sum_{k\in{\mcK^*}}\lambda_k\big|\bfX_{:,G_k}\DDelta_{G_k}\big|_2.
 \end{align}
This readily implies that
$$
\sum_{k\in{\mcK^*}^c}\lambda_k\big|\bfX_{:,G_k}\DDelta_{G_k}\big|_2\leq\sum_{k\in{\mcK^*}}\lambda_k\big|\bfX_{:,G_k}\DDelta_{G_k}\big|_2.
$$
Applying GRE$(\kappa,s)$ assumption and the Cauchy-Schwarz inequality, we get
\begin{align}
\label{trois}
\sum_{k=1}^K\lambda_k\big|\bfX_{:,G_k}\DDelta_{G_k}\big|_2 &\leq 2\Big(\sum_{k\in{\mcK^*}}\lambda_k^2\Big)^{1/2}
\Big(\sum_{k\in{\mcK^*}}\big|\bfX_{:,G_k}\DDelta_{G_k}\big|_2^2\Big)^{1/2}
\leq\frac{2}{\kappa}\big(\sum_{k\in{\mcK^*}}\nolimits\lambda_k^2\big)^{1/2}\;\big|\bfX\DDelta\big|_2.
 \end{align}
It is clear that
\begin{align}
  \big| \bfX\DDelta \big| ^2_2&=\DDelta^\top\bfX^\top(\bfX\widehat{\pphi}-\bfX\tilde{\pphi})\nonumber\\
     &=\DDelta^\top\bfX^\top(\bfX\widehat{\pphi}- \text{\rm diag}(\bfR\widehat{\aalpha})\bY )+\DDelta^\top\bfX^\top(\text{\rm diag}(\bfR\tilde{\aalpha})\bY- \bfX\tilde{\pphi})+ \DDelta^\top\bfX^\top\text{\rm diag}(\bY)\bfR  (\widehat{\aalpha}-\tilde{\aalpha}).\nonumber
\end{align}
In addition, using the relation $\bfX\DDelta=\sum_{k=1}^K\bfX_{:,G_k}\DDelta_{G_k}=\sum_{k=1}^K \Ppi_{G_k}\bfX_{:,G_k}\DDelta_{G_k}$ and the fact that both $( \widehat{\pphi},\widehat{\aalpha})$ and $( \tilde{\pphi},\tilde{\aalpha})$ satisfy constraint (\ref{eq:9}), we have
\begin{align}
 \label{un}
  \big| \bfX\DDelta \big| ^2_2
      &\leq\sum_{k=1}^K\DDelta^\top_{G_k}\bfX^\top_{:,G_k}\Ppi_{G_k}(\bfX\widehat{\pphi}- \text{\rm diag}(\bfR\widehat{\aalpha})\bY )
         +\sum_{k=1}^K\DDelta^\top_{G_k}\bfX^\top_{:,G_k}\Ppi_{G_k}(\text{\rm diag}(\bfR\tilde{\aalpha})\bY- \bfX\tilde{\pphi})\nonumber\\
      &\qquad\qquad+\DDelta^\top\bfX^\top\text{\rm diag}(\bY)\bfR  (\widehat{\aalpha}-\tilde{\aalpha})\nonumber\\
      &\leq2\sum_{k=1}^K\lambda_k\big|\bfX_{:,G_k}\DDelta_{G_k}\big|_2+\big|\bfX\DDelta\big|_2.\big|\bfD_{\bY}\bfR(\widehat{\aalpha}-\tilde{\aalpha}) \big|_2 .
\end{align}
Therefore, from \eqref{trois},
%\begin{align*}
$  | \bfX\DDelta | _2
       \leq\frac{4}{\kappa}\big({\sum_{k\in{\mcK^*}}\nolimits \lambda_k^2}\big)^{1/2} +|\bfD_{\bY}\bfR(\widehat{\aalpha}-\tilde{\aalpha}) |_2
$ %\end{align*}
and we  easily get
\begin{align*}
| \bfX(\widehat{\pphi}-\pphi^*) | _2&\leq  | \bfX(\tilde{\pphi}-\pphi^*) | _2+ | \bfX\DDelta | _2\leq {(z-1)}| \bfX\pphi^* | _2+\frac{4}{\kappa}\Big({\sum_{k\in{\mcK^*}}\nolimits \lambda_k^2}\Big)^{1/2} +  |\bfD_{\bY}\bfR(\widehat{\aalpha}-\tilde{\aalpha}) |_2.
 \end{align*}
where we have used the following notation: for any vector $\bv$, we
denote by $\bfD_\bv$ the diagonal matrix $\diag(\bv)$.

To complete the proof, it suffices to replace $z$ and $\lambda_k$ by their expressions and to use the inequality
\begin{align*}
|\bfD_{\bY}\bfR(\widehat{\aalpha}-\tilde{\aalpha}) |_2
		&\le |\bfD_{\bY}\bfR(\widehat{\aalpha}-\aalpha^*) |_2+(z-1)|\bfD_{\bY}\bfR\aalpha^*|_2\\
		&\le |\bfD_{\bY}\bfR(\widehat{\aalpha}-\aalpha^*) |_2+(z-1)|\bfX\pphi^*+\xxi|_2\\
		&\le |\bfD_{\bY}\bfR(\widehat{\aalpha}-\aalpha^*) |_2+(z-1)\big(|\bfX\pphi^*|_2+|\xxi|_2\big).		
\end{align*}

%%%%%%%%%%%%%%%%%%%
\subsection{Proof of Lemma~\ref{lm1}}
\label{prooflm1}
%%%%%%%%%%%%%%%%%%%

For all $\varepsilon\in(0,1)$, consider the random event $\mathcal{B}_\varepsilon=\bigcap_{\ell=1}^q\left(\mathcal{B}_{\varepsilon,\ell}^2\cap \mathcal{B}_{\varepsilon,\ell}^1\right)$, where
$$\mathcal{B}_{\varepsilon,\ell}^2= \left\{     \sum_{t\in\mcT}\nolimits \frac{r_{t\ell}}{\bR_{t,:}\aalpha^*}\bX_{t,:}\pphi^*\xi_t
\geq-\sqrt{2C_2T\log(2q/\varepsilon)}   \right  \},$$
$$\mathcal{B}_{\varepsilon,\ell}^1=  \left\{     \sum_{t\in\mcT}\nolimits \frac{r_{t\ell}}{\bR_{t,:}\aalpha^*} (\xi_t^2-1)\geq-2\sqrt{C_1T\log(2q/\varepsilon)}      \right  \} . $$

Using standard tail estimates for the Gaussian and the $\chi^2$ distributions, in conjunction with the union bound,
one easily checks that $P(\mathcal{B}_\varepsilon)\geq1-\varepsilon$. In what follows, we show that on the event $\mcB_\varepsilon$,
 $(\tilde{\pphi},\tilde{\aalpha},\tilde{\bv})$ satisfies constraints (\ref{eq:9})-(\ref{eq:11}).

Constraints (\ref{eq:11}) are satisfied (with equality) by definition of $\tilde\bv$. To check that (\ref{eq:10}) is satisfied as well,
we should verify that for all $\ell=1,\ldots, q$,
$$
\frac{1}{z^2}\sum_{t\in\mcT}\nolimits \frac{r_{t\ell}}{\bR_{t,:}\aalpha^*}\le \sum_{t\in\mcT}\nolimits \frac{r_{t\ell}}{\bR_{t,:}\aalpha^*}\bX_{t,:}\pphi^*\xi_t+   \sum_{t\in\mcT}\nolimits \frac{r_{t\ell}}{\bR_{t,:}\aalpha^*}\xi_t^2.
$$
On the event $\mathcal{B}_\varepsilon$, the right-hand side of the last inequality can be lower bounded as follows:
$$
 \sum_{t\in\mcT}\nolimits \frac{r_{t\ell}}{\bR_{t,:}\aalpha^*}\bX_{t,:}\pphi^*\xi_t+   \sum_{t\in\mcT}\nolimits \frac{r_{t\ell}}{\bR_{t,:}\aalpha^*}\xi_t^2\geq -(\sqrt{C_2}+\sqrt{2C_1})\sqrt{2T\log(2q/\varepsilon)} +\sum_{t\in\mcT}\nolimits \frac{r_{t\ell}}{\bR_{t,:}\aalpha^*}.
 $$
Thus, on  $\mathcal{B}_\varepsilon$ if for all $\ell=1,\ldots, q$ %the constraint (\ref{eq:11}) is fulfilled by  $(\tilde{\pphi},\tilde{\aalpha})$ if for all $\ell=1,\ldots, q$
%$$
%\frac{1}{z^2}\sum_{t\in\mcT}\nolimits \frac{r_{t\ell}}{\bR_{t,:}\aalpha^*}\le -\sqrt{2T\log(2q/\varepsilon)}(C_2+\sqrt{2}C_1) +\sum_{t\in\mcT}\nolimits \frac{r_{t\ell}}{\bR_{t,:}\aalpha^*}.
% $$
\begin{align}
\label{C1}
\frac{z^2-1}{z^2}\sum_{t\in\mcT}\nolimits \frac{r_{t\ell}}{\bR_{t,:}\aalpha^*}&\geq (\sqrt{C_2}+\sqrt{2C_1})\sqrt{2T\log(2q/\varepsilon)}
\end{align}
then constraint (\ref{eq:11}) is fulfilled by  $(\tilde{\pphi},\tilde{\aalpha},\tilde\bv)$. Inequality \eqref{C1}
is valid since for any $z\geq 1$
\begin{align*}
\frac{z^2-1}{z^2}\sum_{t\in\mcT}\nolimits \frac{r_{t\ell}}{\bR_{t,:}\aalpha^*} =
\frac{z-1}{z}\bigg(1+\frac1z\bigg)\sum_{t\in\mcT}\nolimits \frac{r_{t\ell}}{\bR_{t,:}\aalpha^*}
&\geq \frac{z-1}{z} TC_3
\end{align*}
and $\frac{z-1}{z} TC_3 \ge (\sqrt{C_2}+\sqrt{2C_1})\sqrt{2T\log(2q/\varepsilon)}$ when $z=1+2C_4\sqrt{\frac{2\log(2q/\varepsilon)}{T}}\le 2$.

On the other hand, since $z\le 2$, a sufficient condition implying that the pair $(\tilde{\pphi},\tilde{\aalpha})$ satisfies (\ref{eq:9}) is
\begin{align}
\label{C2}
2\big|\Ppi_{G_k}\xxi\big|_2&\le {\lambda_k},\qquad \forall k\in\{1,\ldots,K\}.
\end{align}
Recall that $r_k$ denotes  the rank of  $\Ppi_{G_k}$.
Let $\mathcal{R}_\varepsilon$ be the random event of probability at least $1-\varepsilon$ defined as follows
$$
\mathcal{R}_\varepsilon=\bigcap_{k=1}^K\mathcal{R}_{\varepsilon,k}=\bigcap_{k=1}^K\left\{  \big|\Ppi_{G_k}\xxi|_2^2\leq r_k+2\sqrt{  r_k \log(K/\varepsilon)}+2\log(K/\varepsilon)
 \right  \}.
 $$
To prove that $P(\mathcal{R}_\varepsilon)\geq1-\varepsilon $, we use the fact that $\Big|\Ppi_{G_k}\xxi|_2^2$ is drawn from
the $\chi^2_{r_k}$ distribution. Using well-known tail bounds for the $\chi^2$
distribution, we get
%. It is obvious $r_k\leq |G_k|$. For all $k=1,\ldots,K$, consider $\eta_1,\ldots,\eta_{r_k} $, $r_k$  i.i.d centered gaussian variables of variance 1 then%such that $\sum_{i=1}^{r_k}\eta_i^2$ and $ \Big|\Ppi_{G_k}\xxi|_2$ are equal in law, we get let us define $r_k$,  the rank of  $|\Ppi_{G_k}|$
$P(\mathcal{R}_{\varepsilon,k} ^c)\leq \frac{\varepsilon}{K}$. Then, we conclude by the union bound.

Since we chose
$$
 2(r_k+2\sqrt{  r_k \log(K/\varepsilon)}+2\log(K/\varepsilon))^{1/2}={\lambda_k},
$$
on the event  $\mathcal{R}_{\varepsilon}$ inequality \eqref{C2} is satisfied by $(\tilde{\pphi},\tilde{\aalpha})$.

Finally, the triplet $(\tilde{\pphi},\tilde{\aalpha},\tilde\bv)$ fulfills constraints (\ref{eq:9})-(\ref{eq:11}) on the
event $\mathcal{B}_\varepsilon\cap\mathcal{R}_\varepsilon$, which is of a probability at least $1-2\varepsilon$.

%%%%%%%%%%%%%%%%%%%
\subsection{Proof of Theorem~\ref{th3}}
%%%%%%%%%%%%%%%%%%%

We start by noting that, the ScHeDs $(\hat\pphi,\hat\aalpha)$ satisfies $\forall \ell\in\{1,\ldots,q\}$, the relation
\begin{align}
 \sum_{t\in\mcT}\nolimits \frac{r_{t\ell}}{\bR_{t,:}\hat\aalpha}&= \sum_{t\in\mcT}\nolimits  \big( y_t\bR_{t,:}\hat\aalpha -\bX_{t,:}\hat\pphi\big)y_tr_{t\ell}.\label{eq:saturated}
\end{align}
% In what follows, for a vector $\bv$, we will
% denote by $\bfD_\bv$ the diagonal matrix $\diag(\bv)$.
First, for the ScHeDs, all the inequalities in (\ref{eq:11}) are equalities. Indeed, $v_t$'s are only involved in (\ref{eq:10}) and (\ref{eq:11}) and
if we decrease one $v_t$ to achieve an equality in (\ref{eq:11}), the left-hand side of (\ref{eq:10}) will decrease as well and the
constraint will stay inviolated. Thus, setting $\widehat v_t=1/\bR_{t,:}\hat\aalpha$, we get from (\ref{eq:10})
\begin{align}
\sum_{t\in\mcT}\nolimits \frac{r_{t\ell}}{\bR_{t,:}\hat\aalpha}\le \sum_{t\in\mcT}\nolimits {\big( y_t\bR_{t,:}\hat\aalpha -\bX_{t,:}\hat\pphi\big)y_tr_{t\ell}},
\qquad \forall \ell\in\{1,\ldots,q\}.\label{eq:unsaturated}
\end{align}
To be convinced that Eq.\ \eqref{eq:saturated} is true, let us consider for simplicity the one dimensional case $q=1$.
If inequality (\ref{eq:unsaturated}) was strict, for some $w\in(0,1)$, the pair $(w\widehat{\pphi},w\widehat{\aalpha})$ would also satisfy all the constraints of the ScHeDs and the
corresponding penalty term would be smaller than that of $(\widehat{\pphi},\widehat{\aalpha})$. This is impossible since $\widehat{\pphi}$ is an optimal solution.
Thus we get
\begin{align}
\label{eq:11b}
 \sum_{t\in\mcT}\nolimits {\bR_{t,:}^\top}({\bR_{t,:}\hat\aalpha})^{-1} &= \sum_{t\in\mcT}\nolimits  {\bR_{t,:}^\top y_t\big( y_t\bR_{t,:}\hat\aalpha -\bX_{t,:}\hat\pphi\big)}
    = \bfR^\top\bfD_\bY\big(\bfD_\bY\bfR\hat\aalpha-\bfX\hat\pphi\big).
\end{align}
Using the identity $(\bR_{t,:}\hat\aalpha)^{-1} = (\bR_{t,:}\aalpha^*)^{-1} + (\bR_{t,:}\hat\aalpha\bR_{t,:}\aalpha^*)^{-1} \bR_{t,:}(\aalpha^*-\hat\aalpha)$, we get
\begin{align}
\label{eq:12b}
\Big[ \sum_{t\in\mcT} \frac1{(\bR_{t,:}\hat\aalpha)(\bR_{t,:}\aalpha^*)} \bR_{t,:}^\top\bR_{t,:}\Big](\aalpha^*-\hat\aalpha)
    &=-\sum_{t\in\mcT} \frac1{\bR_{t,:}\aalpha^*} \bR_{t,:}^\top+\bfR^\top\bfD_\bY\big(\bfD_\bY\bfR\hat\aalpha-\bfX\hat\pphi\big)\nonumber\\
    &=-\bfR^\top\bfD_{\bfR\aalpha^*}^{-1}\1_T+\bfR^\top\bfD_\bY^2\bfR(\hat\aalpha-\aalpha^*)-\bfR^\top\bfD_\bY\bfX(\hat\pphi-\pphi^*)\nonumber\\
    &\qquad +\bfR^\top\bfD_\bY\big(\bfD_{\bY}\bfR\aalpha^*-\bfX\pphi^*\big).
\end{align}
In view of the identities $\bfD_{\bY}\bfR\aalpha^*-\bfX\pphi^*=\xxi$ and $\bfD_{\bY}=\bfD_{\bfR\aalpha^*}^{-1}(\bfD_{\bfX\pphi^*}+\bfD_{\xxi})$, Eq.\ (\ref{eq:12b})
yields\footnote{We denote by $\xxi^2$ the vector $(\xi_t^2)_{t\in\mcT}$.}
\begin{align}
\label{eq:13b}
\bfR^\top\Big[\bfD_{\bY}^2+\bfD_{\bfR\aalpha^*}^{-1}\bfD_{\bfR\hat\aalpha}^{-1}\Big]\bfR(\aalpha^*-\hat\aalpha)
    =\bfR^\top\bfD_{\bfR\aalpha^*}^{-1}(\xxi^2-\1_T)-\bfR^\top\bfD_{\bY}\bfX(\hat\pphi-\pphi^*)+\bfR^\top\bfD_{\bfR\aalpha^*}^{-1}\bfD_{\bfX\pphi^*}\xxi.
\end{align}
As a consequence, denoting by $\bfM$ the Moore-Penrose pseudo-inverse of the matrix
$\bfR^\top\big[\bfD_{\bY}^2+\bfD_{\bfR\aalpha^*}^{-1}\bfD_{\bfR\hat\aalpha}^{-1}\big]\bfR$,
\begin{align}
\label{eq:14b}
\bfR(\aalpha^*-\hat\aalpha)
    =\bfR\bfM\bfR^\top\bigg(\bfD_{\bfR\aalpha^*}^{-1}(\xxi^2-\1_T)-\bfD_{\bY}\bfX(\hat\pphi-\pphi^*)+\bfD_{\bfR\aalpha^*}^{-1}\bfD_{\bfX\pphi^*}\xxi\bigg).
\end{align}
Multiplying both sides by $\bfD_\bY$ and taking the Euclidean norm, we get
\begin{align}
\label{eq:15b}
\big|\bfD_\bY\bfR(\aalpha^*-\hat\aalpha)\big|_2
    &\le \Big|\bfD_\bY\bfR\bfM\bfR^\top\Big(\bfD_{\bfR\aalpha^*}^{-1}(\xxi^2-\1_T)+\bfD_{\bfR\aalpha^*}^{-1}\bfD_{\bfX\pphi^*}\xxi\Big)\Big|_2
        +\Big|\bfD_\bY\bfR\bfM\bfR^\top\bfD_{\bY}\bfX(\pphi^*-\hat\pphi)\Big|_2.
\end{align}
At this stage of the proof, the conceptual part is finished and we enter into the technical part. At a heuristic level, the first norm
in the right-hand side of (\ref{eq:15b}) is bounded in probability while the second norm is bounded from above by $(1-c)\big|\bfX(\pphi^*-\hat\pphi)\big|_2$
for some constant $c\in(0,1)$. Let us first state these results formally, by postponing their proof to the next subsection, and to finalize the
proof of the theorem.

\begin{lemma}\label{lem:3}
Let $q$ and $T$ be two integers such that $1\le q\le T$ and let $\varepsilon\in(0,1/3)$ be some constant. Assume that for some constant ${\hat D_1}\ge 1$ the inequality $\max_{t\in\mcT}\frac{\bR_{t,:}\hat\aalpha}{\bR_{t,:}\aalpha^*}\le {\hat D_1}$ holds true. Then, on an event of probability at least
$1-3\varepsilon$, the following inequalities are true\footnote{Here and in the sequel, the spectral norm of a matrix $\AA$ is denoted by $\matnorm{\AA}$.}:
\begin{align}
&\matnorm{\bfM^{1/2}\bfR^\top\bfD_{\bY}}\le 1, \\
&\Big|\bfM^{1/2}\bfR^\top\big(\bfD_{\bfR\aalpha^*}^{-1}(\xxi^2-\1_T)+\bfD_{\bfR\aalpha^*}^{-1}\bfD_{\bfX\pphi^*}\xxi\big)\Big|_2
    \le 10\sqrt{q{\hat D_1}\log(2T/\varepsilon)\log(2q/\varepsilon)},\label{eq:15c}\\
&\matnorm{\bfD_\bY\bfR\bfM\bfR^\top\bfD_{\bY}} \le 1-\frac1{2{\hat D_1}\big(|\bfX\pphi^*|_\infty^2+|\xxi|_\infty^2\big)+1}
\le 1-\frac1{{\hat D_1}\big(2|\bfX\pphi^*|_\infty^2+5\log(2T/\varepsilon)\big)}.   \label{eq:20b}
\end{align}
\end{lemma}

In view of these bounds, we get that on an event of probability at least $1-3\varepsilon$,
\begin{align}
\label{eq:15e}
\big|\bfD_\bY\bfR(\aalpha^*-\hat\aalpha)\big|_2
    &\le 10\sqrt{q{\hat D_1}\log(2T/\varepsilon)\log(2q/\varepsilon)}+\bigg(1-\frac1{{\hat D_1}\big(2|\bfX\pphi^*|_\infty^2+5\log(2T/\varepsilon)\big)}\bigg)\big|\bfX(\pphi^*-\hat\pphi)\big|_2.
\end{align}
Combining this inequality with Theorem~\ref{th2} and using the inequality $2|\bfX\pphi^*|_2+|\xxi|_2\le \sqrt{T}\big(2|\bfX\pphi^*|_\infty+|\xxi|_\infty\big)$,
we get that the following inequalities are satisfied with probability $\ge 1-5\varepsilon$:
\begin{align}
\big| \bfX(\widehat{\pphi}-\pphi^*) \big| _2& \leq 2C_4{\hat D_1}\big(2|\bfX\pphi^*|_\infty^2+5\log(2T/\varepsilon)\big)\sqrt{{2\log(2q/\varepsilon)}} (2| \bfX\pphi^* |_\infty+|\xxi|_\infty)\nonumber\\
&\qquad + \frac{8}{\kappa}\Big(2 \sparsity^*+3K^* \log(K/\varepsilon)\Big)^{1/2}{\hat D_1}\big(2|\bfX\pphi^*|_\infty^2+5\log(2T/\varepsilon)\big)\nonumber\\
&\qquad + 10{\hat D_1}\big(2|\bfX\pphi^*|_\infty^2+5\log(2T/\varepsilon)\big)\sqrt{q{\hat D_1}\log(2T/\varepsilon)\log(2q/\varepsilon)} \nonumber\\
& \leq 4 C_4{\hat D_1}\big(2|\bfX\pphi^*|_\infty^2+5{\log(2T/\varepsilon)}\big)^{3/2}\sqrt{{2\log(2q/\varepsilon)}}\nonumber\\
&\qquad + \frac{8{\hat D_1}}{\kappa}\big(2|\bfX\pphi^*|_\infty^2+5\log(2T/\varepsilon)\big)\big(2 \sparsity^*+3K^* \log(K/\varepsilon)\big)^{1/2}\nonumber\\
&\qquad + 10{\hat D_1}\big(2|\bfX\pphi^*|_\infty^2+5\log(2T/\varepsilon)\big)\sqrt{q{\hat D_1}\log(2T/\varepsilon)\log(2q/\varepsilon)}.
\end{align}
Using the notation $D_{T,\varepsilon}={\hat D_1}\big(2|\bfX\pphi^*|_\infty^2+5\log(2T/\varepsilon)\big)$, we obtain
\begin{align}
\big| \bfX(\widehat{\pphi}-\pphi^*) \big| _2& \leq
4 C_4D_{T,\varepsilon}^{3/2}\sqrt{{2\log(2q/\varepsilon)}}+ \frac{8D_{T,\varepsilon}}{\kappa}\big(2 \sparsity^*+3K^* \log(K/\varepsilon)\big)^{1/2}\nonumber\\
&\qquad + 10D_{T,\varepsilon}\sqrt{q{\hat D_1}\log(2T/\varepsilon)\log(2q/\varepsilon)}.\label{eq:16c}
\end{align}
To further simplify the last term, we use the inequalities:
\begin{align*}
10D_{T,\varepsilon}\sqrt{q{\hat D_1}\log(2T/\varepsilon)\log(2q/\varepsilon)}&=D_{T,\varepsilon}\sqrt{10}\sqrt{5{\hat D_1}\log(2T/\varepsilon)}\sqrt{2q\log(2q/\varepsilon)}\\
&\le 4 D_{T,\varepsilon}^{3/2}\sqrt{2q\log(2q/\varepsilon)}.
\end{align*}
Combining this with (\ref{eq:16c}) yields (\ref{bound:2}).

To prove (\ref{bound:3}), we use once again (\ref{eq:14b}) to infer that
\begin{align*}
\big|\bfR(\aalpha^*-\hat\aalpha)\big|_2
    &\le \Big|\bfR\bfM\bfR^\top\Big(\bfD_{\bfR\aalpha^*}^{-1}(\xxi^2-\1_T)+\bfD_{\bfR\aalpha^*}^{-1}\bfD_{\bfX\pphi^*}\xxi\Big)\Big|_2
        +\Big|\bfR\bfM\bfR^\top\bfD_{\bY}\bfX(\pphi^*-\hat\pphi)\Big|_2\nonumber\\
    &\le \matnorm{\bfR\bfM^{1/2}}\Big(\Big|\bfM^{1/2}\bfR^\top\Big(\bfD_{\bfR\aalpha^*}^{-1}(\xxi^2-\1_T)+\bfD_{\bfR\aalpha^*}^{-1}\bfD_{\bfX\pphi^*}\xxi\Big)\Big|_2
        +\matnorm{\bfM^{1/2}\bfR^\top\bfD_{\bY}}\Big|\bfX(\pphi^*-\hat\pphi)\Big|_2\Big).
\end{align*}
In view of Lemma~\ref{lem:3}, this leads to
\begin{align}
\label{eq:17b}
\big|\bfR(\aalpha^*-\hat\aalpha)\big|_2
    &\le \matnorm{\bfR\bfM^{1/2}}\Big(10\sqrt{q{\hat D_1}\log(2T/\varepsilon)\log(2q/\varepsilon)}+\Big|\bfX(\pphi^*-\hat\pphi)\Big|_2\Big),
\end{align}
with probability at least $1-5\varepsilon$. Using the bound in (\ref{bound:2}), we get
\begin{align}
\label{eq:18b}
\big|\bfR(\aalpha^*-\hat\aalpha)\big|_2
    &\le \matnorm{\bfR\bfM^{1/2}}\Big(4 (C_4+2)D_{T,\varepsilon}^{3/2}\sqrt{{2q\log(2q/\varepsilon)}}+ \frac{8D_{T,\varepsilon}}{\kappa}\sqrt{2 \sparsity^*+3K^* \log(K/\varepsilon)}\Big).
\end{align}
In view of the inequality\footnote{We use the notation $\AA \succeq\BB$ and $\BB\preceq\AA$ for indicating that the matrix $\AA-\BB$ is positive semi-definite. For any matrix $\AA$,
we denote by $\AA^+$ its Moore-Penrose pseudoinverse.}
\begin{align*}
(\bfR\bfM^{1/2})(\bfR\bfM^{1/2})^\top
		&= \bfR\big[\bfR^\top(\bfD_\bY^2+\bfD_{\bfR\aalpha^*}^{-1}\bfD_{\bfR\hat\aalpha}^{-1})\bfR\big]^+\bfR^\top\\
		&\preceq \bfR\big[\bfR^\top(\bfD_{\bfR\aalpha^*}^{-1}\bfD_{\bfR\hat\aalpha}^{-1})\bfR\big]^+\bfR^\top\\
		&\preceq (\max_{t\in\mcT} [\bR_{t,:}\aalpha^*\cdot\bR_{t,:}\hat\aalpha])\bfR\big[\bfR^\top\bfR\big]^+\bfR^\top		
\end{align*}
we get
\begin{align*}
\matnorm{\bfR\bfM^{1/2}}^2
	&=\matnorm{(\bfR\bfM^{1/2})(\bfR\bfM^{1/2})^\top } \\
	&\le \hat D_1 \big|\bfR\aalpha^*\big|^2_\infty \cdot\matnorm{\bfR\big[\bfR^\top\bfR\big]^+\bfR^\top}\\
	&\le \hat D_1 \big|\bfR\aalpha^*\big|^2_\infty,
\end{align*}
where the last inequality follows from the fact that $\bfR\big[\bfR^\top\bfR\big]^+\bfR^\top$ is an orthogonal projector.
%%%%%%%%%%%%%%%%%%%%%%%%%%%%%%%%%%%%%%%%%%%
\subsection{Proof of Lemma~\ref{lem:3}}
%%%%%%%%%%%%%%%%%%%%%%%%%%%%%%%%%%%%%%%%%%%
We start by presenting a proof of (\ref{eq:15c}). We have
\begin{align}
\label{eq:15d}
\Big|\bfD_\bY\bfR\bfM\bfR^\top\big(\bfD_{\bfR\aalpha^*}^{-1}(\xxi^2&-\1_T)+\bfD_{\bfR\aalpha^*}^{-1}\bfD_{\bfX\pphi^*}\xxi\big)\Big|_2\nonumber\\
    &\le\matnorm{\bfD_\bY\bfR\bfM^{1/2}}\cdot\big|\bfM^{1/2}\bfR^\top\big(\bfD_{\bfR\aalpha^*}^{-1}(\xxi^2-\1_T)+\bfD_{\bfR\aalpha^*}^{-1}\bfD_{\bfX\pphi^*}\xxi\big)\big|_2\nonumber\\
    &\le\matnorm{\bfD_\bY\bfR\bfM^{1/2}}\cdot\big(\big|\bfM^{1/2}\bfR^\top\bfD_{\bfR\aalpha^*}^{-1}(\xxi^2-\1_T)\big|_2
        +\big|\bfM^{1/2}\bfR^\top\bfD_{\bfR\aalpha^*}^{-1}\bfD_{\bfX\pphi^*}\xxi\big|_2\big).
\end{align}
We remark that
$$
\bfM^{+}=\bfR^\top\big[\bfD_{\bY}^2+\bfD_{\bfR\aalpha^*}^{-1}\bfD_{\bfR\hat\aalpha}^{-1}\big]\bfR\succeq \bfR^\top\bfD_{\bY}^2\bfR\quad\Longrightarrow\quad \matnorm{\bfD_{\bY}\bfR\bfM^{1/2}}\le 1.
$$
and that
$$
\bfM^+\succeq  \bigg(\min_t \frac{y_t^2+(\bR_{t,:}\aalpha^*\cdot\bR_{t,:}\hat\aalpha)^{-1}}{(\bX_{t,:}\pphi^*/\bR_{t,:}\aalpha^*)^2}\bigg)
\bfR^\top\bfD_{\bfR\aalpha^*}^{-2}\bfD_{\bfX\pphi^*}^{2}\bfR,
$$
which implies that
\begin{align}
\label{eq:16b}
\big|\bfM^{1/2}\bfR^\top\bfD_{\bfR\aalpha^*}^{-1}\bfD_{\bfX\pphi^*}\xxi\big|_2^2
        &=\xxi^\top\bfD_{\bfR\aalpha^*}^{-1}\bfD_{\bfX\pphi^*} \bfR\bfM\bfR^\top\bfD_{\bfX\pphi^*}\bfD_{\bfR\aalpha^*}^{-1}\xxi\nonumber\\
        &\le \bigg(\max_{t\in\mcT}\frac{(\bX_{t,:}\pphi^*)^2}{(\bR_{t,:}\aalpha^*)^2y_t^2+(\bR_{t,:}\aalpha^*/\bR_{t,:}\hat\aalpha)}\bigg)\xxi^\top\Ppi_1\xxi,
\end{align}
where $\Ppi_1=\bfD_{\bfR\aalpha^*}^{-1}\bfD_{\bfX\pphi^*}\bfR\big(\bfR^\top \bfD_{\bfX\pphi^*}^2\bfD_{\bfR\aalpha^*}^{-2}\bfR\big)^+\bfR^\top \bfD_{\bfX\pphi^*}\bfD_{\bfR\aalpha^*}^{-1}$
is the orthogonal projection on the linear subspace of $\RR^T$ spanned by the columns of the matrix $\bfD_{\bfR\aalpha^*}^{-1}\bfD_{\bfX\pphi^*}\bfR$. By the Cochran theorem, the random variable
$\eta_1=\xxi^\top\Ppi_1\xxi$ is distributed according to the $\chi^2_q$ distribution.

Using similar arguments based on matrix inequalities, one checks that
\begin{align}
\label{eq:17c}
\big|\bfM^{1/2}\bfR^\top\bfD_{\bfR\aalpha^*}^{-1}(\xxi^2-\1_T)\big|_2^2
        &\le \bigg(\max_{t\in\mcT}\frac{(\bR_{t,:}\aalpha^*)^{-2}}{y_t^2+(\bR_{t,:}\aalpha^*\cdot\bR_{t,:}\hat\aalpha)^{-1}}\bigg)(\xxi^2-1)^\top\Ppi_2(\xxi^2-1)\nonumber\\
        &\le \bigg(\max_{t\in\mcT}\frac{\bR_{t,:}\hat\aalpha}{\bR_{t,:}\aalpha^*}\bigg)\underbrace{(\xxi^2-1)^\top\Ppi_2(\xxi^2-1)}_{=:\eta_2},
\end{align}
where $\Ppi_2=\bfD_{\bfR\aalpha^*}^{-1}\bfR\big(\bfR^\top \bfD_{\bfR\aalpha^*}^{-2}\bfR\big)^+\bfR^\top \bfD_{\bfR\aalpha^*}^{-1}$ is the orthogonal projection
on the linear subspace of $\RR^T$ spanned by the columns of the matrix $\bfD_{\bfR\aalpha^*}^{-1}\bfR$.

To further simplify (\ref{eq:16b}), one can remark that under the condition $\bR_{t,:}\hat\aalpha\le {\hat D_1}\bR_{t,:}\aalpha^*$, it holds
\begin{align}
\label{eq:18c}
\frac{(\bX_{t,:}\pphi^*)^2}{(\bR_{t,:}\aalpha^*)^2y_t^2+(\bR_{t,:}\aalpha^*/\bR_{t,:}\hat\aalpha)}
    &\le \frac{(\bX_{t,:}\pphi^*)^2}{(\bX_{t,:}\pphi^*+\xi_t)^2+{\hat D_1}^{-1}} \le {1+{\hat D_1}\xi_t^2}.
\end{align}

These bounds, combined with (\ref{eq:15d}), yield
\begin{align}
\label{eq:19b}
\Big|\bfD_\bY\bfR\bfM\bfR^\top\big(\bfD_{\bfR\aalpha^*}^{-1}(\xxi^2-\1_T)+\bfD_{\bfR\aalpha^*}^{-1}\bfD_{\bfX\pphi^*}\xxi\big)\Big|_2
\le \sqrt{(1+{\hat D_1}|\xxi|_\infty^2)\eta_1}+\sqrt{{\hat D_1}\eta_2}.
\end{align}
One can also notice that $\Ppi_2$ is a projector on a subspace of dimension at most equal to $q$, therefore one can write
$\Ppi_2=\sum_{\ell=1}^q \bv_\ell \bv_\ell^\top$ for some unit vectors $\bv_\ell\in\RR^T$. This implies that
$$
\eta_2 = \sum_{\ell=1}^q |\bv_\ell^\top(\xxi^2-\1_T)|^2\le q \max_{\ell=1,\ldots,q} \Big|\sum_{t\in\mcT} v_{\ell,t} (\xi^2_t-1)\Big|^2.
$$
Hence, large deviations of $\eta_1$ and $\eta_2$ can be controlled using standard tail bounds; see, for instance, \citet[Lemma 1]{LaurentMassart}.
This implies that with probability at least $1-2\varepsilon$,
\begin{align*}
\Big|\bfD_\bY\bfR\bfM\bfR^\top\big(\bfD_{\bfR\aalpha^*}^{-1}(\xxi^2-\1_T)+\bfD_{\bfR\aalpha^*}^{-1}\bfD_{\bfX\pphi^*}\xxi\big)\Big|_2
\le \sqrt{1+{\hat D_1}|\xxi|_\infty^2}(\sqrt{q}+\sqrt{2\log(q/\varepsilon)}) +\sqrt{q {\hat D_1}}\; 4\log(2q/\varepsilon).
\end{align*}
To conclude, it suffices to remark that $\Pb(|\xxi|_\infty \le \sqrt{2\log(2T/\varepsilon)})\ge 1-\varepsilon$. This implies that
\begin{align*}
\Big|\bfD_\bY\bfR\bfM\bfR^\top\big(\bfD_{\bfR\aalpha^*}^{-1}(\bfD_\xxi^2&-\II_T)\1_T+\bfD_{\bfR\aalpha^*}^{-1}\bfD_{\bfX\pphi^*}\xxi\big)\Big|_2\\
    &\le 2\sqrt{{\hat D_1}\log(2T/\varepsilon)}(\sqrt{q}+\sqrt{2\log(q/\varepsilon)}) +\sqrt{q {\hat D_1}}\; 4\log(2q/\varepsilon)\\
    &\le 4\sqrt{2q{\hat D_1}\log(2T/\varepsilon)\log(q/\varepsilon)} +4\sqrt{q {\hat D_1}}\; \log(2q/\varepsilon)\\
    &\le 10\sqrt{q{\hat D_1}\log(2T/\varepsilon)\log(2q/\varepsilon)}.
\end{align*}
This completes the proof of the first claim of the lemma.

Let us now switch to a proof of (\ref{eq:20b}).
It is clear that
\begin{align}
\label{eq:21b}
\matnorm{\bfD_\bY\bfR\bfM\bfR^\top\bfD_{\bY}} &= \matnorm{\bfM^{1/2}\bfR^\top\bfD_{\bY}}^2\nonumber\\
& \le \matnorm{\bfM^{1/2}\bfR^\top(\bfD_{\bY}^2+\bfD_{\bfR\hat\aalpha}^{-1}\bfD_{\bfR\aalpha^*}^{-1})^{1/2}}^2
\matnorm{(\bfD_{\bY}^2+\bfD_{\bfR\hat\aalpha}^{-1}\bfD_{\bfR\aalpha^*}^{-1})^{-1/2}\bfD_\bY}^2\nonumber\\
& \le \matnorm{(\bfD_{\bY}^2+\bfD_{\bfR\hat\aalpha}^{-1}\bfD_{\bfR\aalpha^*}^{-1})^{-1/2}\bfD_\bY}^2\nonumber\\
& = \max_{t\in\mcT} \frac{y_t^2}{y_t^2+(\bR_{t,:}\aalpha^*\cdot\bR_{t,:}\hat\aalpha)^{-1}}.
\end{align}
Using the fact that $\bR_{t,:}\hat\aalpha\le {\hat D_1}\bR_{t,:}\aalpha^*$ for every $t$, we obtain
\begin{align}
\label{eq:22b}
\matnorm{\bfD_\bY\bfR\bfM\bfR^\top\bfD_{\bY}}
 &= \max_{t\in\mcT} \frac{y_t^2(\bR_{t,:}\aalpha^*)^2}{y_t^2(\bR_{t,:}\aalpha^*)^2+{\hat D_1}^{-1}}\nonumber\\
 &= 1-\min_{t\in\mcT} \frac{1}{ {\hat D_1}y_t^2(\bR_{t,:}\aalpha^*)^2+1}\nonumber\\
 &= 1-\min_{t\in\mcT} \frac{1}{ {\hat D_1}(\bX_{t,:}\pphi^*+\xi_t)^2+1}.
\end{align}
To complete the proof of the lemma, it suffices to remark that
$(\bX_{t,:}\pphi^*+\xxi_t)^2 \le 2(\bX_{t,:}\pphi^*)^2+2\xi_t^2\le 2|\bfX\pphi^*|_\infty^2+2|\xxi|_\infty^2$
and to apply the well-known bound on the tails of the Gaussian distribution.

%%%%%%%%%%%%%%%%%%%%

\end{document}